\documentclass[%
 aip,
 amsmath,amssymb,
 reprint,%
]{revtex4-1}

\usepackage{graphicx}
\usepackage{dcolumn}
\usepackage{bm}

\usepackage[utf8]{inputenc}
\usepackage[T1]{fontenc}
\usepackage{mathptmx}
\usepackage{etoolbox}
\usepackage{natbib}
\usepackage{mathtools}
\usepackage{booktabs}
\usepackage{supertabular}
\usepackage{multirow}

\newcommand{\indep}{\perp\!\!\!\perp}
\makeatletter
\def\@email#1#2{%
 \endgroup
 \patchcmd{\titleblock@produce}
  {\frontmatter@RRAPformat}
  {\frontmatter@RRAPformat{\produce@RRAP{*#1\href{mailto:#2}{#2}}}\frontmatter@RRAPformat}
  {}{}
}%
\makeatother
\begin{document}

\preprint{AIP/123-QED}

\title[Complex-Valued LinPDE-GP for Helmholtz]{Operator-Informed Gaussian Processes for Complex Helmholtz Wavefields: From Synthetic Benchmarks to In Vivo Brain Elastography}

\author{Boyuan Deng}%
 \email{bdeng8@jh.edu}
\affiliation{ 
Department of Civil and Systems Engineering, Johns Hopkins University
}%

\author{Kshitiz Upadhyay}%
 \email{kshitizu@umn.edu}
\affiliation{ 
Department of Aerospace Engineering and Mechanics, University of Minnesota
}%

\author{Michael Shields}
 \email{michael.shields@jhu.edu}
\affiliation{
Department of Civil and Systems Engineering, Johns Hopkins University
}%

\date{\today}

\begin{abstract}
    The Helmholtz equation governs time-harmonic wave propagation, and in dissipative media a complex modulus renders its squared wavenumber $\kappa^2$ complex. Inferring such fields from sparse, noisy data calls for solvers that also quantify their own uncertainty. Physics-informed Gaussian-process (GP) regression supplies this by returning a posterior over the solution, yet operator-conditioned formulations have been developed almost exclusively for real-valued fields. We extend operator-informed GP regression to complex-valued Helmholtz problems by realifying the complex operator into an equivalent coupled real block, which enables inference with standard real-valued GP conditioning. The construction admits a family of priors, from a proper diagonal prior to coregionalized and multiscale variants, and conditions on PDE residuals and boundary traces. On benchmark problems in one to three dimensions, the solver is competitive with finite-difference and neural-network baselines at a far smaller interior-constraint budget. Unlike those deterministic baselines, it returns a posterior over the complex wavefield rather than a point estimate. Applied to \textit{in vivo} brain magnetic resonance elastography, a proper multiscale prior reconstructs the shear curl field to a correlation of $0.77$ with measurement, above a $0.75$ target. The gain arises from the multiscale kernel rather than from real--imaginary coupling. We further identify a low-frequency accuracy ceiling set by model mismatch and a posterior uncertainty that is not yet calibrated. Calibrated uncertainty therefore emerges as the central next step for probabilistic wavefield inference in dissipative media.
\end{abstract}

\maketitle


\section{Introduction}

The Helmholtz equation governs time-harmonic wave phenomena across science and engineering, unifying the frequency-domain description of acoustic and electromagnetic fields~\citep{larsson2003helmholtz, gumerov2005fast}. Its solutions underpin both forward simulation and inverse modeling pipelines, where wave fields are inferred from sparse, noisy measurements or embedded as submodules within larger scientific workflows~\citep{lahaye2017modern, GRYAZIN2003122}. Yet, despite its apparent simplicity, the Helmholtz operator induces numerical challenges that become acute in the high-frequency regime: the solution is highly oscillatory, the operator is indefinite, and accuracy requirements tighten as the wavelength decreases. Classical discretization-based solvers, the Finite Difference Method (FDM), the Finite Element Method (FEM), and related weighted-residual schemes, remain the workhorses of computational wave physics, but they can suffer from dispersion-driven errors and the well-known pollution effect, which forces mesh refinement beyond what is suggested by local approximation order alone~\citep{ihlenburg1995finite, babuvska1995generalized, li2024optimized}. At scale, classical Helmholtz solvers often rely on specialized preconditioning and domain decomposition for large sparse discretizations~\citep{erlangga2008advances, farhat2000two, diwan2021iterative}, and when dense operators arise, as in boundary integral formulations, on hierarchical or low-rank compression to control memory and runtime~\citep{cheng2006wideband, martinsson2005fast}.

These numerical difficulties are compounded by epistemic uncertainty that is intrinsic to many contemporary applications. Material parameters (and sometimes sources or boundary data) are often uncertain, spatially heterogeneous, and inferred indirectly from sparse, noisy measurements; moreover, in dissipative or viscoelastic media, attenuation is commonly modeled through complex-valued constitutive laws, yielding complex propagation constants whose real part controls phase advance and whose imaginary part controls decay~\citep{zhao2018time, holm2014comparison}. Such complex-valued formulations arise, for example, when wave propagation is governed by a frequency-dependent complex modulus, producing attenuation that requires complex propagation constants rather than purely real wavenumbers~\citep{carcione1988wave, kjartansson1979constant}. In these contexts, it is not enough to compute a single best-guess solution: downstream decisions often require calibrated uncertainty quantification (UQ) and principled propagation of parameter and data uncertainty through the partial differential equation (PDE) solver.

Physics-informed machine learning has been extensively studied as a complement to classical PDE solvers and, in some settings, as a meshfree surrogate. Physics-informed neural networks (PINNs) are a prominent example, encoding governing equations by augmenting the training loss with PDE-residual and boundary/initial-condition terms~\citep{raissi2019physics, mao2020physics, jagtap2020conservative, kharazmi2021hp, karniadakis2021physics, cai2021physics}. However, vanilla PINN-style approaches ultimately deliver point estimates whose uncertainty quantification is indirect, sensitive to optimization and regularization choices, and not inherently tied to discretization error~\citep{yang2021b, cuomo2022scientific, linka2022bayesian, zou2025uncertainty}. Probabilistic numerical methods help address this gap by casting numerical computation as a statistical inference problem and returning distributions that quantify solver (discretization) error. When propagated through PDE-constrained Bayesian inverse problems, this can mitigate overconfident posterior concentration and yield more conservative inference~\citep{hennig2015probabilistic, cockayne2017probabilistic, abdulle2021probabilistic}. In this context, physics-informed Gaussian process (GP) regression offers an attractive solution for many problems. Because GPs are closed under linear operators, PDE constraints and boundary observations can be incorporated as linear information, yielding posterior solution distributions with coherent uncertainty estimates~\citep{bishop2006pattern, Rasmussen2006Gaussian}. In computational mechanics, this closure underlies Gaussian functional regression for linear PDEs~\citep{nguyen2015gaussian}, GP regression constrained by boundary value problems~\citep{gulian2022gaussian}, the statistical finite element method~\citep{girolami2021statistical}, and finite-element GPs~\citep{dalton2026finite}. Recent work has formalized this perspective by showing that interpreting linear PDE solvers as physics-informed GP regression strictly generalizes methods of weighted residuals (MWRs, including collocation, finite volumes, pseudospectral, and Galerkin methods) and equips them with a structured error estimate and inbuilt model-parameter uncertainty propagation~\citep{pförtner2024physicsinformedgaussianprocessregression}.

Nevertheless, a practical limitation remains for frequency-domain wave problems: much of the physics-informed GP literature and constructions based on conditioning on linear operators are developed for real-valued fields~\citep{pförtner2024physicsinformedgaussianprocessregression, albert2019gaussian, nguyen2015gaussian, gulian2022gaussian, dalton2026finite}. Separately from PDE-oriented physics-informed GPs, complex-valued GP regression has been developed primarily in signal processing and time series analysis, where proper/improper (noncircular) complex signals are modeled via covariance and pseudo-covariance~\citep{ambrogioni2019complex, boloix2018complex, 7178363}. This gap persists even in PDE-tailored kernel methods that explicitly treat the Helmholtz equation, where complex-valued fields are commonly identified as an open direction rather than a supported capability~\citep{albert2019gaussian}. Uncertainty-aware treatments of the Helmholtz equation itself remain rare. Recent examples are parameterized PINNs for interior acoustics~\citep{mao2025decomposition} and probabilistic operator learning at high frequency~\citep{zou2026probabilistic}, neither of which is a Gaussian-process formulation. The absence of complex-valued support is in sharp contrast to signal processing, where complex-valued regression is grounded in a mature proper/improper framework~\citep{neeser2002proper, schreier2010statistical}. Bridging these lines of work is not a cosmetic generalization. Complex-valued Helmholtz problems require inference machinery that respects both the operator structure of the PDE and the joint covariance of the real–imaginary representation.

Here we introduce a complex-field extension of operator-informed, physics-informed GP regression for Helmholtz-type wave problems. We make three contributions:
\begin{enumerate}
    \item We extend the LinPDE-GP framework to complex-valued fields through a real–imaginary (real2) construction, in which a complex-coefficient linear operator is realified into an equivalent coupled 2×2 real block. This allows complex PDEs to be solved with standard real-valued GP conditioning and transpose-based adjoints, with PDE residuals and boundary traces incorporated as linear information operators. We further develop a family of admissible real2 priors. These span a proper (circular) diagonal prior, an intrinsic coregionalization model (ICM), and a multiscale linear model of coregionalization (LMC) that combines length scales and can couple the real and imaginary channels.
    \item We instantiate the framework for the Helmholtz operator with complex-valued squared wavenumber $\kappa^2$, covering the dissipative regime in which a complex modulus gives $\operatorname{Im}(\kappa^2)\neq 0$. The resulting probabilistic solver returns a posterior over the complex wavefield together with a structured estimate of the error arising from finite operator information. On closed-form problems in one, two, and three dimensions, it attains accuracy comparable to the FDM and PINNs while conditioning on substantially fewer interior constraints.
    \item We demonstrate the solver on \textit{in vivo} human-brain magnetic resonance elastography (MRE) data, reconstructing the curl wavefield of a variable-coefficient Helmholtz problem whose spatially varying $\kappa^2(\mathbf{x})$ is fixed from independently inverted modulus maps. The multiscale LMC prior clears the acceptance target ($\mathrm{Pearson}>0.75$) where a single-scale prior does not. An ablation isolates the multiscale base kernel, rather than real–imaginary coupling, as the source of the gain. We also delineate the boundaries of the method on real data, reporting a low-frequency ceiling set by physical model mismatch and a posterior standard deviation that tracks that mismatch only weakly.
\end{enumerate}

\section{\label{sec:formu_n_notation} Problem formulation}

Let $\Omega \subset \mathbb{R}^d$ be a domain with boundary $\partial\Omega$. We study the inhomogeneous Helmholtz equation as follows:
\begin{equation}
    (\Delta + \kappa^2)u=f\quad\text{in }\Omega,\qquad u=g\ \text{on }\partial\Omega,
    \label{eqn:helmholtz}
\end{equation}
where $\Delta$ is the Laplace operator, $u:\Omega\to\mathbb{C}$ is the field of interest (e.g., pressure or displacement), $\mathbb{C}$ denotes the set of complex numbers, $\kappa\in\mathbb{C}$ is the wavenumber, and $f\in H^{-1}(\Omega)$ is a source term. In our convention, the source term appears with a positive sign on the right-hand side; this choice reflects a purely conventional assignment of sign (as some formulations instead adopt the form $(-\Delta - \kappa^2)u=f$).

Physically, $\kappa^2$ encapsulates both material and frequency effects. For instance, in the case of linear, time-harmonic shear waves propagating through a linear viscoelastic medium of density $\rho$, angular frequency $\omega$, and complex shear modulus $G' + \mathrm{i}G''$ ($G'$: Storage modulus; $G''$: Loss modulus), it can be expressed as
\begin{equation}
    \kappa^2 = \frac{\rho \omega^2}{G' + \mathrm{i}G''}
\end{equation}
following the sign convention of~\citet{meyers2008mechanical}.


This formulation is generalized in two ways later in the paper. In heterogeneous media, the shear modulus, and hence the squared wavenumber, varies in space. The coefficient then becomes $\kappa^2=\kappa^2(\mathbf{x})$, and the governing equation holds pointwise. This variable-coefficient form underlies the \textit{in vivo} brain reconstruction of Sec.~\ref{exp:brain}. The second generalization concerns vector wavefields such as those measured in elastography. For a measured displacement field $\mathbf{u}$, we condition not on $\mathbf{u}$ but on the curl observable $\mathbf{q}=\nabla\times\mathbf{u}$, which removes the irrotational (compressional) component. Under a local-homogeneity assumption, each Cartesian component of $\mathbf{q}$ satisfies a scalar Helmholtz equation of the same form. The scalar treatment developed here therefore applies to each component independently.

A complete list of notation used throughout the manuscript can be found in Appendix~\ref{app:notation}.

\section{Methodology}

\subsection{Gaussian Processes for Real and Complex Fields}

\subsubsection{Real-Valued Gaussian Process Regression}

Let $\mathcal{X}$ denote the input space. A Gaussian process (GP) $\{h(x)\}_{x\in\mathcal{X}}$ comprises a collection of random variables indexed by $x\in\mathcal{X}$, for which every finite subset follows a joint Gaussian distribution~\citep{Rasmussen2006Gaussian}.
A (real-valued) GP  is fully specified by its mean function $m(x) = \mathbb{E}[h(x)]$ and covariance function (or kernel) $k(x, x') = \mathbb{E}[(h(x)-m(x))(h(x')-m(x'))]$. The kernel function encodes structural assumptions about the latent function, such as smoothness, stationarity, or periodicity. We denote this as $h(x) \sim \mathcal{GP}(m, k)$.

In a standard regression setting, we observe a dataset $\{(x_i, y_i)\}_{i=1}^N$, where the targets $y_i$ are corrupted by zero-mean independent Gaussian noise $\varepsilon_i \sim \mathcal{N}(0, \sigma_{\varepsilon}^2)$, such that $y_i = h(x_i) + \varepsilon_i$. For a set of training inputs $X$ and test inputs $X_*$, the joint distribution of the observed targets $\mathbf{y}$ and the latent function values $\mathbf{h}_*$ is given by:
\begin{equation}
    \begin{bmatrix} \mathbf{y} \\ \mathbf{h}_* \end{bmatrix} \sim \mathcal{N}\left( \begin{bmatrix} m(X) \\ m(X_*) \end{bmatrix}, \begin{bmatrix} K(X,X) + \sigma_{\varepsilon}^2 I_N & K(X,X_*) \\ K(X_*,X) & K(X_*,X_*) \end{bmatrix} \right),
\end{equation}
where $K(X, X')$ denotes the matrix of covariances evaluated at all pairs of points in $X$ and $X'$~\citep{bishop2006pattern}.

By conditioning the joint prior on the observations, we obtain the posterior distribution over the latent function values at the test inputs, $p(\mathbf{h}_* | X, \mathbf{y}, X_*) = \mathcal{N}(\bar{\mathbf{h}}_*, \mathbb{V}[\mathbf{h}_*])$. The predictive mean and covariance are derived as:
\begin{align}
    \bar{\mathbf{h}}_* &= m(X_*) + K(X_*, X) [K(X, X) + \sigma_{\varepsilon}^2 I_N]^{-1} (\mathbf{y} - m(X)),
    \label{eqn:mean_cov1}
    \\
    \mathbb{V}[\mathbf{h}_*] &= K(X_*, X_*) - K(X_*, X) [K(X, X) + \sigma_{\varepsilon}^2 I_N]^{-1} K(X, X_*).
    \label{eqn:mean_cov2}
\end{align}
Here the symmetry of the kernel, $k(x,x') = k(x',x)$, implies $K(X_*,X) = K(X,X_*)^\top$, so the cross-covariance blocks in Eqs.~\eqref{eqn:mean_cov1}--\eqref{eqn:mean_cov2} are transposes of one another, and the Gram matrix $K(X,X)$ is symmetric. The vector $\alpha = [K(X, X) + \sigma_{\varepsilon}^2 I_N]^{-1} (\mathbf{y} - m(X))$ can be interpreted as a set of dual coefficients, which weight the kernel evaluations $K(X_*, X)$ to form the predictive mean.

In the physics-informed setting, the same formulas apply when the observations are generalized to linear functionals of the latent function $h$ (e.g., residuals of linear PDE operators and boundary traces), as used later in this work.

\subsubsection{Complex-Valued Gaussian Process Regression}

For the complex-valued extension, let $h(x) = h_R(x) + \mathrm{i} h_I(x)$, where $h_R$ and $h_I$ form a jointly Gaussian two-output real-valued process. A complex-valued Gaussian process is fully specified by its mean, its covariance function $k(x,x') = \mathbb{E}\big[(h(x)-m(x))\overline{(h(x')-m(x'))}\big]$ which satisfies the Hermitian symmetry $k(x,x') = \overline{k(x',x)}$, and its pseudo-covariance function $c(x,x') = \mathbb{E}\big[(h(x)-m(x))(h(x')-m(x'))\big]$.

In many machine learning treatments, one assumes a proper complex GP (also known as circularly symmetric in the zero-mean case), meaning the pseudo-covariance vanishes ($c(x,x') = 0$)~\citep{neeser2002proper}. For a proper complex GP, the real and imaginary parts are jointly Gaussian, pointwise uncorrelated, and have equal marginal variance. Consequently, 
$h_R(x)$ and $h_I(x)$ are independent random variables at any fixed location $x$. Under this assumption, the complex multivariate Gaussian density for a vector $\mathbf{h} \in \mathbb{C}^N$ is given by~\citep{neeser2002proper, schreier2010statistical}:
\begin{equation}
    p(\mathbf{h}) = \frac{1}{\pi^N \det(K)} \exp\left( -(\mathbf{h} - \mathbf{m})^H K^{-1} (\mathbf{h} - \mathbf{m}) \right).
\end{equation}
Here, $K$ is the Hermitian positive-definite covariance matrix with entries $K_{ij}=k(x_i,x_j)$, and we write $\mathbf{h} \sim \mathcal{CN}(\mathbf{m},K)$, where $\mathcal{CN}$ denotes a proper complex Gaussian distribution with mean $\mathbf{m}$, covariance $K$, and zero pseudo-covariance. The operator $(\cdot)^H$ denotes the Hermitian (conjugate) transpose. Under the properness assumption for both the prior and the (Gaussian) observation noise, and assuming independence between them, the regression equations for the predictive mean and covariance retain the algebraic structure of the real-valued case in Eqs.~\eqref{eqn:mean_cov1} -- \eqref{eqn:mean_cov2}, with the transposes induced by kernel symmetry replaced by Hermitian transposes: the Hermitian symmetry $k(x,x')=\overline{k(x',x)}$ gives $K(X_*,X)=K(X,X_*)^{H}$ in place of $K(X,X_*)^{\top}$, and the Gram matrix $K(X,X)$ is Hermitian rather than symmetric~\citep{8307269}. In practice, we represent the complex-valued GP as an equivalent two-output real-valued GP over the real and imaginary parts of $\mathbf{h}$.

\subsection{Physics-Informed GP framework (LinPDE-GP)}

In this work, we adopt and extend the probabilistic framework introduced by~\citet{pförtner2024physicsinformedgaussianprocessregression}, referred to as LinPDE-GP, which recasts the solution of linear PDEs as Bayesian inference. Unlike classical methods of weighted residuals that produce a single deterministic estimate, LinPDE-GP places a problem-specific GP prior over the solution space and conditions it on physical constraints, including the PDE operator, boundary conditions, and noisy measurements, by utilizing bounded linear operators. This formulation yields a posterior GP with closed-form mean and covariance functions, providing both a point estimate that recovers classical methods of weighted residuals and a structured quantification of epistemic uncertainty. While the original framework was derived for real-valued fields, the Helmholtz equation necessitates a treatment of complex amplitudes and phases. To accommodate this, we organize the development in two steps: first, we briefly recapitulate the foundational real-valued LinPDE-GP formulation to establish notation and the conditioning framework; second, we extend this formulation to the complex domain via an isomorphism between $\mathbb{C}$ and $\mathbb{R}^2$ (referred to as the real2 representation). This mapping enables the direct application of real-valued GP inference machinery to complex wave physics, specifically the Helmholtz equation considered in this paper.

\subsubsection{Real-Valued LinPDE-GP Formulation}

We briefly recapitulate the foundational real-valued LinPDE-GP framework introduced by~\citet{pförtner2024physicsinformedgaussianprocessregression}. This overview serves to establish the necessary context for our complex-valued extension, but does not present the approach in all its detail. 
For a comprehensive treatment of the original methodology, we refer the reader to the primary publication. 
Our focus in the subsequent subsections is specifically directed toward the application of this framework to the Helmholtz equation, detailing the construction of the requisite computational modules and experimental validation.

\citet{pförtner2024physicsinformedgaussianprocessregression} view the solution of a linear (strong or weak) boundary value problem as Bayesian inference in a GP model, where all physics constraints and data enter through \emph{affine bounded linear information operators}. Concretely, let $u$ denote the unknown (possibly vector-valued) solution, modeled as a random element in a Banach space $\mathbb{B}$ (or solution space $\mathbb{U}$) of functions over a domain $\Omega$. A key building block is conditioning a GP prior on finite-dimensional observations of the form
\begin{equation}
\begin{aligned}
    y \;&=\; \mathcal{L}[u] + \varepsilon,\\
    \mathcal{L}&:\mathbb{B}\to \mathbb{R}^n \text{ bounded linear},\\
    \varepsilon&\sim\mathcal{N}(\mu,\Sigma),\;\varepsilon\indep u.
\end{aligned}
\end{equation}
where $\varepsilon\indep u$ denotes statistical independence of the noise and the latent solution. If $u\sim\mathcal{GP}(m,k)$, then $\mathcal{L}[u]$ is Gaussian and the posterior $u\,|\,y$ remains a GP with closed-form moments:
\begin{equation}
    m^{u|y}(x)=m(x)+\mathcal{L}[k(x,\cdot)]^\top (\mathcal{L}k\mathcal{L}^\top+\Sigma)^+ \bigl(y-(\mathcal{L}[m]+\mu)\bigr),
\end{equation}

\begin{equation}
    k^{u|y}(x,x')=k(x,x')-\mathcal{L}[k(x,\cdot)]^\top (\mathcal{L}k\mathcal{L}^\top+\Sigma)^+ \mathcal{L}[k(\cdot,x')].
\end{equation}
where $(\cdot)^+$ denotes the Moore--Penrose pseudo-inverse.
This theorem enables a unified treatment of PDE constraints, boundary/initial conditions, and noisy measurements by stacking the corresponding information operators into a single $\mathcal{L}$.

To connect with classical PDE discretizations, the paper introduces \emph{MWR information operators}. Writing the (weak or strong) operator equation as $\mathcal{D}^{(w)}[u]=f^{(w)}$, choose a trial space $\hat{\mathbb{U}}\subseteq \mathbb{U}$ and a bounded (possibly oblique) projection $\mathcal{P}_{\hat{\mathbb{U}}}:\mathbb{U}\to \hat{\mathbb{U}}$. For a test functional $l$ in the appropriate dual space, 
\begin{equation}
    \mathcal{R}_{l,\mathcal{P}_{\hat{\mathbb{U}}}}[(u,f^{(w)})]
    \;:=\;
    (l\circ \mathcal{D}^{(w)}\circ \mathcal{P}_{\hat{\mathbb{U}}})[u] - l[f^{(w)}],
\end{equation}
where $\circ$ denotes composition of operators and functionals, so that $\mathcal{P}_{\hat{\mathbb{U}}}$ is applied first, followed by $\mathcal{D}^{(w)}$, and then by $l$. One then conditions on a finite collection $\{\mathcal{R}_{l^{(i)},\mathcal{P}_{\hat{\mathbb{U}}}}=0\}_{i=1}^n$. By selecting $\hat{\mathbb{U}}=\mathrm{span}\{\phi^{(1)},\dots,\phi^{(M)}\}$ and choosing test functionals as in standard MWRs (collocation, Galerkin, FEM, etc.), the resulting posterior mean becomes directly comparable to traditional solvers.

Finally, \citet{pförtner2024physicsinformedgaussianprocessregression} (Algorithm 1) casts LinPDE-GP as an iterative inference procedure: a policy selects new test functionals, projections/trial spaces, and (optional) measurement weightings. These define a new stacked information operator, after which the Gram matrix and representer weights are updated to produce the next posterior GP over the unknowns.

\subsubsection{Complex-Valued Extension of LinPDE-GP (Real2 Representation)}
\label{subsec:real2}

\begin{figure*}[t]
    \centering
    \includegraphics[width=0.95\textwidth]{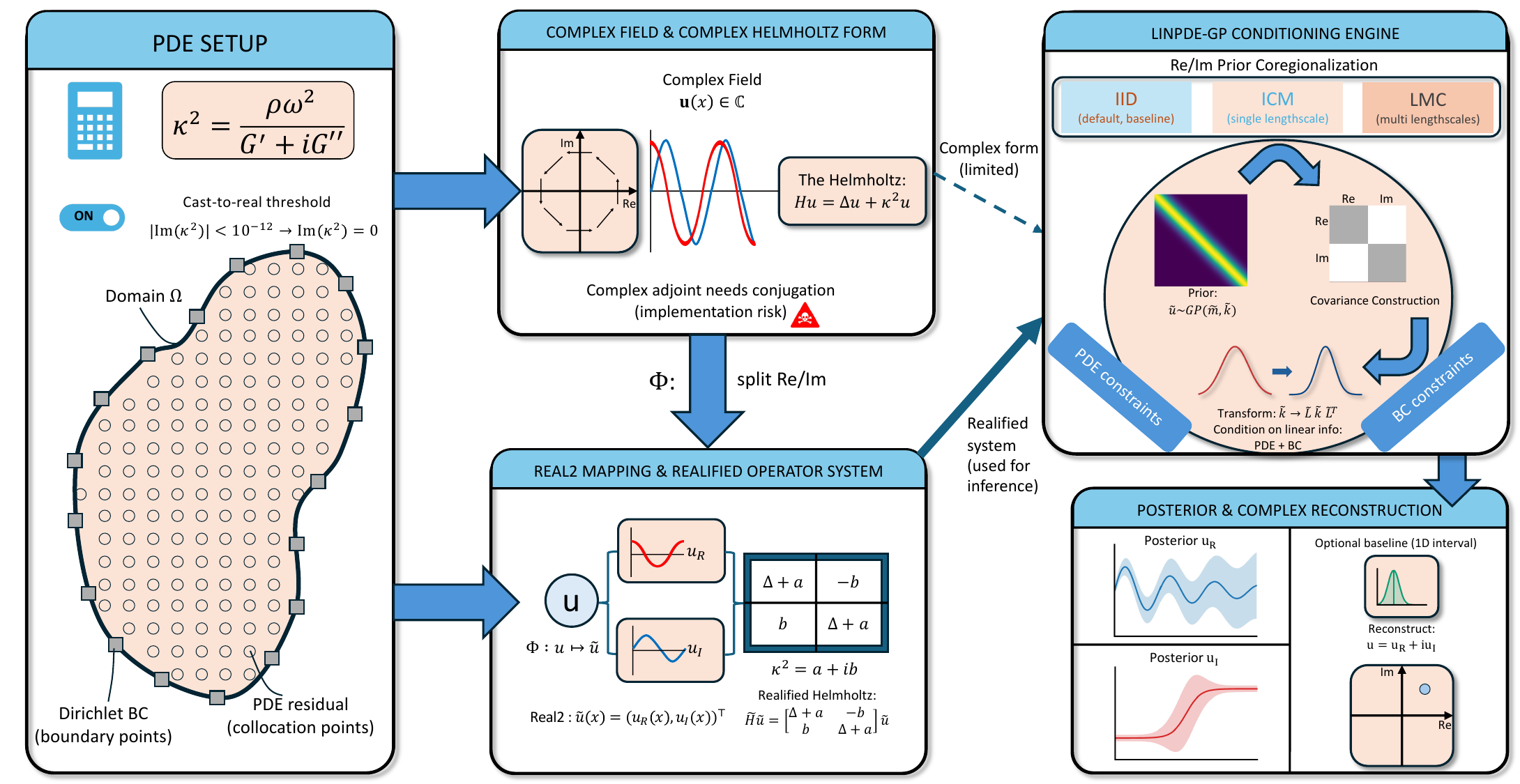} 
    \caption{\label{fig:real2} \textbf{LinPDE-GP architecture for complex wavefield reconstruction.} The framework represents the complex-valued solution $u : \Omega \to \mathbb{C}$ in a "Real2" (real–imaginary) form $\tilde u=(u_R,u_I)^\top$. This isomorphism transforms the Helmholtz operator with complex squared wavenumber $\kappa^2$ into an equivalent coupled $2 \times 2$ real-valued block system. Crucially, this formulation permits conditioning via standard real-valued GP machinery (using transpose adjoints $\mathcal{L}^{\top}$) while bypassing the implementation complexity of complex Hermitian adjoints. The inference engine by default instantiates a block-diagonal prior, treating real and imaginary components as independent channels, conditioned on PDE residuals and Dirichlet boundaries to yield the posterior reconstruction $u(x) = u_R(x) + \mathrm{i}\,u_I(x)$. More general coregionalized priors relax this channel independence and combine several length scales.}
\end{figure*}

Fig.~\ref{fig:real2} gives an overview of the construction developed in this subsection. We extend the LinPDE-GP framework to support complex-valued PDEs, specifically Helmholtz-type problems, by representing complex fields via a real-valued vector isomorphism with the existing real-valued GP machinery and the linearity of differential operators. In the remainder of this work, we denote by $u : \Omega \to \mathbb{C}$ the complex-valued PDE solution field, which plays the role of the latent function $h$ introduced in the preceding subsections. We define the real two-dimensional (hereafter, "real2") representation of $u$ as the $\mathbb{R}^2$-valued field $\tilde{u} : \Omega \to \mathbb{R}^2$ given by
\begin{equation}
\begin{aligned}
u_R(x) &:= \operatorname{Re} (u(x)), \qquad u_I(x) := \operatorname{Im} (u(x)),\\
\tilde{u}(x) &=
\begin{pmatrix}
u_R(x)\\
u_I(x)
\end{pmatrix}.
\end{aligned}
\end{equation}
The mapping
\begin{equation}
    \Phi : u \mapsto \tilde{u}
\end{equation}
is an $\mathbb{R}$-linear isomorphism between the vector space of complex-valued functions on $\Omega$ and the space of $\mathbb{R}^2$-valued functions on $\Omega$. Its inverse reconstructs the complex field from its components,
\begin{equation}
    u(x) = u_R(x) + \mathrm{i}\,u_I(x),
\end{equation}
where $u_R$ and $u_I$ denote the real and imaginary parts of $\tilde{u}$, respectively.

Having defined $\Phi : u \mapsto \tilde{u}$ and its inverse $u(x) = u_R(x) + \mathrm{i}\,u_I(x)$, we formulate the probabilistic model and inference in the real2 representation. Concretely, we place a GP prior directly on the $\mathbb{R}^2$-valued field $\tilde u:\Omega\to\mathbb{R}^2$, and we only recover complex-valued quantities $u$ through the inverse mapping when needed for interpretation or output.

We therefore consider a vector-valued GP
\begin{equation}
    \tilde{u} \sim \mathcal{GP}(\tilde{m}, \tilde{k}),
    \quad
    \tilde{m}:\Omega\to\mathbb{R}^2,
    \quad
    \tilde{k}:\Omega\times\Omega\to\mathbb{R}^{2\times 2},
\end{equation}
with mean
\begin{equation}
    \tilde{m}(x)=
    \begin{pmatrix}
    \operatorname{Re} (m(x))\\
    \operatorname{Im} (m(x))
    \end{pmatrix},
\end{equation}
and matrix-valued covariance
\begin{equation}
\tilde{k}(x, x') =
\begin{pmatrix}
k_{\mathrm{RR}}(x, x') & k_{\mathrm{RI}}(x, x') \\
k_{\mathrm{IR}}(x, x') & k_{\mathrm{II}}(x, x')
\end{pmatrix},
\end{equation}
where the blocks are ordinary real covariances between components, e.g.
\begin{equation}
\begin{split}
    k_{\mathrm{RR}}(x,x') &= \operatorname{Cov}\bigl( u_R(x), u_R(x')\bigr) \\
    k_{\mathrm{RI}}(x,x') &=\operatorname{Cov}\bigl( u_R(x), u_I(x')\bigr)
\end{split}
\end{equation}
and analogously for $k_{\mathrm{IR}}$ and $k_{\mathrm{II}}$. As usual for matrix-valued kernels, admissibility of $\tilde k$ means that it is symmetric in the sense
\begin{equation}
    \tilde{k}(x,x')=\tilde{k}(x',x)^{\top}
\end{equation}
and that its block Gram matrices are positive semidefinite for any finite set of input locations.

We organize the real2 prior into a baseline diagonal construction and two coregionalized generalizations, all of which are admissible choices of the matrix-valued kernel $\tilde{k}$ defined above.

In the present work, we instantiate this real2 prior using the diagonal “proper/circular” construction implemented in the codebase: \url{https://github.com/bydeng01/linpde-gp}

We choose a single real-valued scalar kernel
\begin{equation}
    k : \Omega \times \Omega \to \mathbb{R}
\end{equation}
and set
\begin{equation}
\begin{split}
    k_{\mathrm{RR}}(x,x') &= k_{\mathrm{II}}(x,x') = \tfrac{1}{2} k(x,x'), \\
    k_{\mathrm{RI}}(x,x') &= k_{\mathrm{IR}}(x,x') = 0.
\end{split}
\end{equation}
so that
\begin{equation}
\tilde{k}(x,x') = \frac{1}{2} k(x,x')
\begin{pmatrix}
1 & 0 \\
0 & 1
\end{pmatrix}.
\end{equation}
Under the inverse mapping $u = u_R + \mathrm{i} u_I$, this yields a complex-valued field whose real and imaginary parts are independent real GPs with identical covariance $k/2$; in particular, the induced complex covariance satisfies
\begin{equation}
\mathbb{E}\bigl[(u(x) - m(x)) \overline{(u(x') - m(x'))}\bigr]= k(x,x'),
\end{equation}
and variance $\operatorname{Var}\bigl(u(x)\bigr) = k(x,x)$. This diagonal construction is the simplest admissible real2 prior: a single length scale with no coupling between channels. We now introduce two generalizations that the brain experiments of Sec.~\ref{exp:brain} compare against it.

The \emph{intrinsic coregionalization model}~\cite{goovaerts1997geostatistics,alvarez2012kernels} (ICM) pairs a single base kernel $k$ with a symmetric positive-semidefinite coregionalization matrix $B\in\mathbb{R}^{2\times2}$ acting on the $(u_R,u_I)$ channels,
\begin{equation}
    \tilde{k}(x,x') = k(x,x')\,B,
    \qquad
    B = \frac{1}{2}\begin{pmatrix} 1 & \rho_{\mathrm{RI}} \\ \rho_{\mathrm{RI}} & 1 \end{pmatrix},
    \quad \rho_{\mathrm{RI}}\in(-1,1).
\end{equation}
Setting $\rho_{\mathrm{RI}}=0$ recovers the diagonal proper prior above. For $\rho_{\mathrm{RI}}\neq0$ the off-diagonal blocks are non-zero and symmetric, $k_{\mathrm{RI}}=k_{\mathrm{IR}}=\tfrac12\rho_{\mathrm{RI}}\,k$. This violates the properness condition $k_{\mathrm{RI}}=-k_{\mathrm{IR}}$, so the resulting prior is improper (noncircular) and couples the two channels a priori.

The \emph{linear model of coregionalization}~\cite{journel1976mining,alvarez2012kernels} (LMC) sums $Q$ such terms, each with its own base kernel and coregionalization matrix,
\begin{equation}
    \tilde{k}(x,x') = \sum_{q=1}^{Q} k_q(x,x')\,B_q,
    \qquad B_q \succeq 0.
\end{equation}
The ICM is the $Q=1$ case, and the diagonal proper prior is $Q=1$ with $B_1=\tfrac12 I_2$. Choosing the base kernels $k_q$ at distinct length scales $\ell_q$ produces a multiscale prior. Taking $B_q=\tfrac12\sigma_q^2 I_2$ in every term (that is, $\rho_{\mathrm{RI}}=0$ throughout) gives
\begin{equation}
    \tilde{k}(x,x') = \frac{1}{2}\Big(\textstyle\sum_{q=1}^{Q}\sigma_q^2\,k_q(x,x')\Big)\, I_2,
\end{equation}
a proper prior whose base kernel is an additive combination of length scales. This isolates two effects that we separate empirically in Sec.~\ref{exp:brain}. The multiscale structure of the base kernel is retained while the prior stays proper, whereas the cross-channel coupling $\rho_{\mathrm{RI}}$ requires leaving the proper family.

Linear operators act on complex-valued GPs through the same real2 identification. Let $\mathcal{L}$ be a $\mathbb{C}$-linear operator acting on complex-valued functions on $\Omega$ (for example, a linear differential operator arising from a PDE), and write
\begin{equation}
(\mathcal{L}u)(x) = \mathcal{L}_x u(x).
\end{equation}
We apply operators to the real2 GP and interpret the result as complex via the inverse mapping when needed. Under the real2 isomorphism, we represent $\mathcal{L}$ as a $2 \times 2$ block operator acting on $\tilde{u}$,
\begin{equation}
\tilde{\mathcal{L}} =
\begin{pmatrix}
\mathcal{L}_{11} & \mathcal{L}_{12} \\
\mathcal{L}_{21} & \mathcal{L}_{22}
\end{pmatrix},
\end{equation}
where each block $\mathcal{L}_{ij}$ acts linearly on real-valued functions. Due to the $\mathbb{C}$-linearity of $\mathcal{L}$, these blocks satisfy the constraints $\mathcal{L}_{11} = \mathcal{L}_{22}$ and $\mathcal{L}_{21} = -\mathcal{L}_{12}$.

Any such $\mathbb{C}$-linear operator admits a unique decomposition of this form, obtained by applying $\mathcal{L}$ to the real and imaginary parts separately and collecting real and imaginary components; equivalently,
\begin{equation}
    \mathcal{L}(u_R + \mathrm{i}u_I)
    = \bigl(\mathcal{L}_{11}u_R + \mathcal{L}_{12}u_I\bigr)
      + \mathrm{i}\bigl(\mathcal{L}_{21}u_R + \mathcal{L}_{22}u_I\bigr),
\end{equation}
for real-valued functions $u_R$ and $u_I$.

The transformed real2 field $\tilde{\mathcal{L}}\tilde{u}$ is a vector-valued GP with mean and covariance given by the standard linear-Gaussian transformation rules:
\begin{equation}
\tilde{m}_{\tilde{\mathcal{L}}\tilde{u}}(x) = \tilde{\mathcal{L}}_x \tilde{m}(x),
\end{equation}
and
\begin{equation}
\tilde{k}_{\tilde{\mathcal{L}}\tilde{u}}(x,x')
= \tilde{\mathcal{L}}_x \,\tilde{k}(x,x')\, \tilde{\mathcal{L}}_{x'}^{\top},
\end{equation}
where $\tilde{\mathcal{L}}_x$ acts on the first argument and $\tilde{\mathcal{L}}_{x'}^{\top}$ denotes the operator transpose acting on the second. This expression follows directly from the joint Gaussianity of any finite collection of evaluations of $u$ and the linearity of $\mathcal{L}$.

In summary, the real2 representation allows us to apply the LinPDE-GP framework to complex-valued fields using standard real-valued GP machinery. In this work, we instantiate this construction for proper complex GPs derived from a single real kernel, for coregionalized priors that couple the real and imaginary channels and combine multiple length scales, and for Helmholtz-type operators.

\subsection{Helmholtz Instantiation}

We instantiate the time-harmonic Helmholtz operator in a form motivated by shear-wave propagation in a linear viscoelastic solid,
\begin{equation}
    \mathcal{H}u=\Delta u + \kappa^2u,
    \qquad
    \kappa^2=\frac{\rho\omega^2}{G' + \mathrm{i}G''},
\end{equation}
where $\Delta$ is the spatial Laplacian and the (possibly complex) squared wavenumber $\kappa^2\in\mathbb{C}$ encodes both elastic storage and viscous dissipation through the storage and loss moduli $(G',G'')$.

Equivalently, writing $\kappa^2=a+\mathrm{i}b$,
\begin{equation}
\begin{aligned}
a &= \operatorname{Re}(\kappa^2)
   = \frac{\rho\omega^2 G'}{(G')^2+(G'')^2}, \\
b &= \operatorname{Im}(\kappa^2)
   = -\frac{\rho\omega^2 G''}{(G')^2+(G'')^2}.
\end{aligned}
\end{equation}

The treatment depends on whether the squared wavenumber is real or complex. As a limiting special case, for $\kappa^2\in\mathbb{R}$, the problem reduces to a scalar real-valued Helmholtz equation, which standard real-valued GP conditioning handles directly. The same problem may equivalently be embedded in the two-output real--imaginary form of Sec.~\ref{subsec:real2}, in which case the realified operator is block-diagonal. For the complex-valued case considered here, the real and imaginary channels are intrinsically coupled, and we therefore work in the real--imaginary representation throughout. The realification yields the coupled operator
\begin{equation}
    \tilde{\mathcal{H}}=
    \begin{pmatrix}
        \Delta + a & -b\\
        b & \Delta + a
    \end{pmatrix},
    \qquad
    \tilde{\mathcal{H}}\tilde u = \widetilde{\mathcal{H}u},
\end{equation}
which is algebraically equivalent to applying $\mathcal{H}$ to $u$ and separating the real and imaginary parts. In this block representation, taking the transpose corresponds to conjugating $\kappa^2$ under the real–imaginary identification (equivalently, $b\mapsto -b$).

Probabilistic inference is performed by conditioning a GP prior on linear information derived from the PDE and boundary conditions. Dirichlet boundary conditions are represented by an identity linear function operator on each boundary component. Upon choosing boundary evaluation sites, this yields linear observation functionals (point-evaluation composed with the boundary operator). Interior physics is enforced by residual constraints $(\mathcal{H}u)(x)=f(x)$ (default $f\equiv 0$) at collocation sites, implemented as point-evaluation composed with the differential operator. To support this realification, we construct a two-output prior from a scalar kernel $k$. Unless stated otherwise, we use the proper diagonal prior, with independent real and imaginary channels and block-diagonal covariance
\begin{equation*}
    \mathrm{Cov}\big[(u_R,u_I)(x),(u_R,u_I)(x')\big] = \tfrac{1}{2}k(x,x')\,I_2,
\end{equation*}
so that each component has marginal covariance $k/2$ and there is no cross-covariance between $u_R$ and $u_I$ in the prior. The ICM and LMC priors generalize this choice through a coregionalization matrix and a sum of base kernels; the diagonal form is the $B=\tfrac12 I_2$, single-scale special case.

Where analytic solutions are available, we optionally attach closed-form baselines on one-dimensional interval domains via strict type-based checks, and therefore only when the domain is provided in an explicit interval form. In particular, on interval domains, the trivial solution is attached strictly when the Dirichlet data are specified as an identically zero boundary function (numerically zero constants are not detected by this check), without additionally verifying that $f\equiv 0$; it should therefore be interpreted as a convenience baseline for the homogeneous case. Otherwise, when the forcing is absent or identically zero as recognized by the implementation, the complex-exponential family $u(x)=c_1 e^{\mathrm{i} \kappa x}+c_2 e^{-\mathrm{i} \kappa x}$ is attempted and used when the associated coefficient system is numerically nonsingular.

As a numerical convention, when $\lvert\operatorname{Im}(\kappa^2)\rvert < 10^{-12}$ we treat the
coefficient as real. This decouples the real and imaginary blocks of the realified operator and removes spurious cross-channel coupling driven purely by floating-point roundoff, at the cost of neglecting damping below this scale.

\section{\label{sec:exp_n_results} Benchmark Experiments and Results}

\paragraph{Baseline Comparisons.} We benchmark against the FDM and PINN, representing classical mesh-based and learning-based solvers, respectively. Because baseline implementations and tuning are experiment-dependent, we report the configuration used in each setting within the corresponding subsection. Because the baselines are deterministic, cross-method comparisons necessarily focus on point-estimate error metrics. A distinguishing feature of LinPDE-GP is that conditioning yields a full posterior distribution over the solution field, providing a structured estimate of the approximation uncertainty as an integral part of inference rather than requiring additional post-hoc analysis. We report posterior uncertainty throughout to illustrate this capability and to diagnose where the imposed information set most and least constrains the solution.

\paragraph{Error metrics.} Cross-method comparisons rest on four point-estimate metrics that quantify the discrepancy between a solver's estimate $\hat{u}$ and the reference solution $u^*$. All are evaluated on a test grid of $N_{\text{test}}$ points, specified per experiment. Let $e_i = \hat{u}_i - u^*_i$ denote the pointwise error, with $i$ indexing the test points. We report the mean squared error, $\mathrm{MSE} = \tfrac{1}{N_{\text{test}}}\sum_{i} |e_i|^2$; the mean absolute error, $\mathrm{MAE} = \tfrac{1}{N_{\text{test}}}\sum_{i} |e_i|$; the maximum absolute error, $\max_i |e_i|$; and the relative $\ell^2$ error, $\lVert \hat{u}-u^* \rVert_2 / \lVert u^* \rVert_2$. For the complex-valued problems, all four metrics are computed separately for the real and imaginary components $u_R$ and $u_I$. The definitions are identical across LinPDE-GP and the FDM and PINN baselines, so the reported errors are directly comparable; for LinPDE-GP, $\hat{u}$ is the posterior mean.

\paragraph{Conditioning information.} Across experiments, LinPDE-GP conditions on (i) boundary observations that enforce the prescribed boundary values on $\partial\Omega$, and (ii) PDE Observations that constrain the operator application $\mathcal{L}u$ to match the forcing $f$ at interior collocation points. Unless stated otherwise, observations are treated as noiseless, and collocation points are placed uniformly within a fixed interior sub-interval to avoid boundary clustering. We vary the number of collocation points to study accuracy, data trade-offs, and comparisons with deterministic baselines.

\paragraph{Analytical reference solutions.} The two- and three-dimensional benchmarks share a common eigenfunction-series representation. For a constant forcing $f$ with homogeneous Dirichlet data on $\Omega=[-1,1]^{d}$, separation of variables gives the sine eigenfunction expansion
\begin{equation}
\begin{aligned}
    u(\mathbf{x}) &= \sum_{\substack{k_1,\dots,k_d\ge 1\\ k_1,\dots,k_d\ \mathrm{odd}}}
    \Big(\tfrac{4}{\pi}\Big)^{d}\frac{f}{k_1\cdots k_d}\,
    \frac{\displaystyle\prod_{j=1}^{d}\sin\frac{k_j\pi(x_j+1)}{2}}{-\lambda_{\mathbf{k}}+\kappa^{2}},
    \\
    \lambda_{\mathbf{k}}&=\frac{\pi^{2}}{4}\sum_{j=1}^{d}k_j^{2},
    \label{eqn:nd_analytic}
\end{aligned}
\end{equation}
where the complex squared wavenumber enters solely through the modal denominator $(-\lambda_{\mathbf{k}}+\kappa^{2})^{-1}$. Only odd multi-indices contribute, since even modes have vanishing projection onto the constant source. For the real forcing used here, the imaginary part of $u$ is generated entirely by $\operatorname{Im}(\kappa^{2})$, vanishing when $\kappa^{2}$ is real. The two-dimensional case is $d=2$ (indices $m,n$; prefactor $16f/(mn\pi^{2})$) and the three-dimensional case is $d=3$ (indices $m,n,p$; prefactor $64f/(mnp\pi^{3})$). The one-dimensional benchmarks instead use the elementary closed forms given in their respective subsections. Reported errors use the per-experiment truncations stated below.

\subsection{Validation on the 1D real-valued Helmholtz equation}

We begin with a one-dimensional Helmholtz problem with a real-valued wavenumber, serving as a controlled setting to validate conditioning from boundary data to PDE information. We instantiate the formulation in Sec.\ref{sec:formu_n_notation} on $\Omega=[-1.0,1.0]$ with homogeneous Dirichlet boundary conditions, $u(-1)=u(1)=0$. We set $\kappa=1$ and use a constant forcing $f(x)=2$, for which the solution is available in closed form, $u^*(x)=2\left(1-\cos(x)/\cos(1)\right)$ and is used as ground truth. We place a zero-mean GP prior on $u$ with an exponentiated quadratic kernel, fixing the length-scale $\ell=1.0$ and the kernel amplitude $\sigma_f=2.0$.

\begin{figure*}[t]
    \centering
    \includegraphics[width=0.8\textwidth]{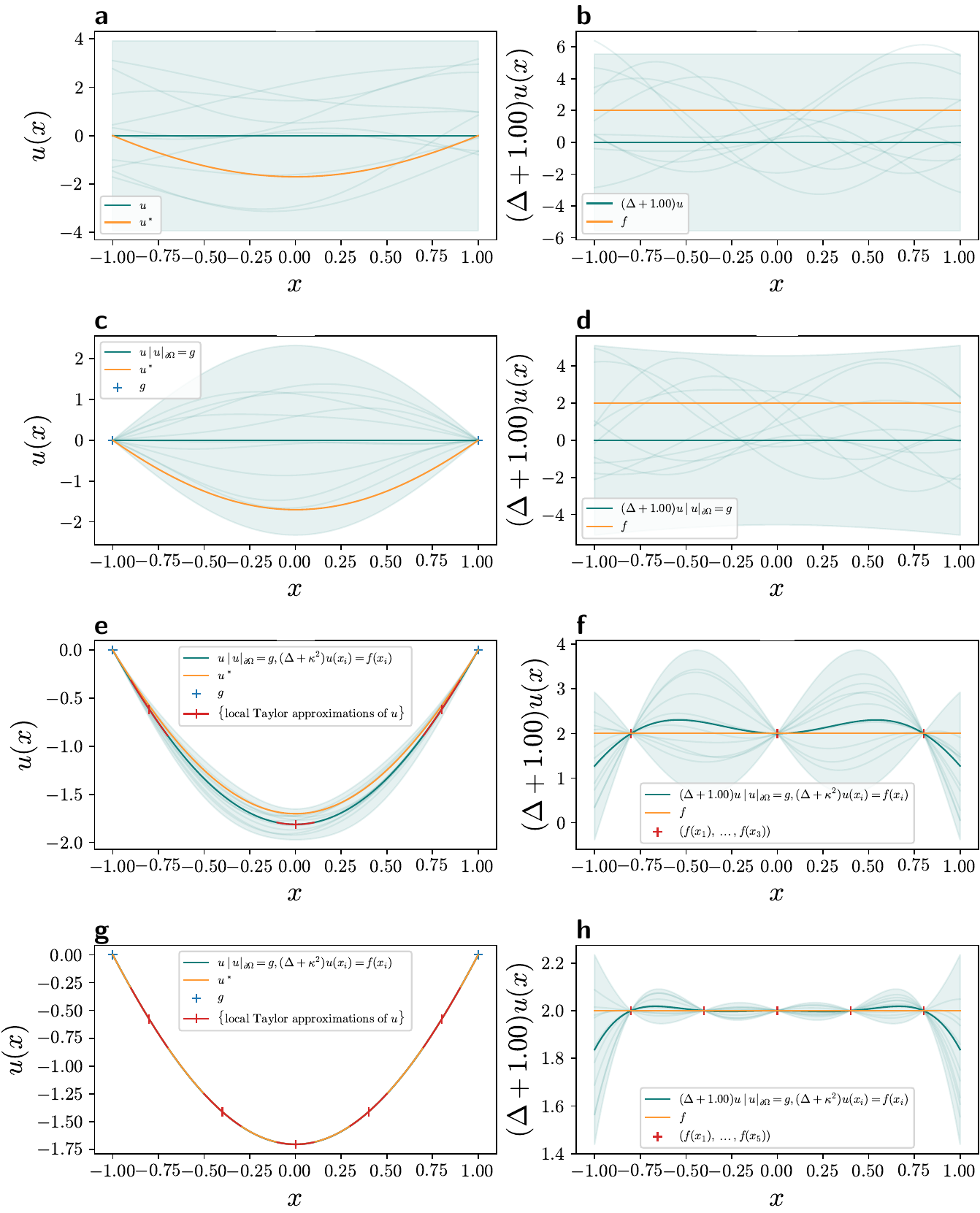}
    \caption{\label{fig:1d_real_3pts} \textbf{1D Real Helmholtz Equation: Posterior evolution under sequential conditioning.} Left column: belief about the solution $u(x)$. Right column: belief about the operator-applied field $(\Delta+\kappa^{2})u(x)$, overlaid with the source term $f$. Rows: (a,b) prior; (c,d) conditioned on the boundary points $g$ at $\partial\Omega$; (e,f) conditioned on the boundary points and $N_{\text{col}}=3$ PDE observations; (g,h) conditioned on the boundary points and $N_{\text{col}}=5$ PDE observations. Blue/teal lines: GP means for $u$ and $(\Delta+\kappa^{2})u$. Shaded bands: 95\% credible intervals. Faint lines: posterior samples. Orange lines: analytical solution $u^*$ and source term $f$. Markers indicate observed components: boundary values $g$ in panels (c), (e), and (g), and collocation values $f(x_i)$ in panels (f) and (h). The short parabolic segments in (e) and (g) visualize the local Taylor approximation of the posterior mean at the collocation sites.}
\end{figure*}

This experiment is designed to (i) visualize how boundary and PDE information enter the posterior through sequential conditioning, and (ii) quantify how accuracy improves as the number of PDE collocation points $N_{\text{col}}$ increases. We therefore report both posterior evolution (Fig.~\ref{fig:1d_real_3pts}) and point-estimate errors against deterministic baselines (Table~\ref{tab:tab1d_real}).

Fig.~\ref{fig:1d_real_3pts} traces the posterior under sequential conditioning. It starts from the prior in (a,b), adds the boundary constraints alone in (c,d), and then adds interior collocation points with $N_{\text{col}}=3$ in (e,f) and $N_{\text{col}}=5$ in (g,h). The $N_{\text{col}}=3$ setting in (e,f) makes the propagation of information and the associated posterior uncertainty easy to interpret. The $N_{\text{col}}=5$ setting in (g,h) then illustrates the further contraction obtained by adding PDE observations.

Conditioning on the boundary observations alone enforces $u(\pm1)=0$ but leaves the interior largely prior-driven, as shown in (c,d). The mismatch between the operator-applied field $(\Delta+\kappa^{2})u$ and the forcing $f$ reflects this, together with wide uncertainty in $(\Delta+\kappa^{2})u$ across the domain. Adding $N_{\text{col}}=3$ PDE observations in (e,f) drives the posterior over $(\Delta+\kappa^{2})u$ toward $f$ at the collocation sites. This information propagates to the primal field $u$ and yields substantial posterior contraction across $\Omega$. Increasing $N_{\text{col}}$ from $3$ to $5$ in (g,h) further sharpens the posterior. Table~\ref{tab:tab1d_real} quantifies the corresponding accuracy gains, showing that adding collocation points lowers all reported error metrics and mirrors the posterior contraction in Fig.~\ref{fig:1d_real_3pts}.

\paragraph{Baseline solvers.}
Table~\ref{tab:tab1d_real} compares the solution with the baseline FDM and PINN. For FDM, we use a standard second-order finite-difference discretization on a uniform grid over $[-1,1]$ with $N_{\text{fdm}}=101$ nodes, enforcing the Dirichlet boundary values $u(-1)=u(1)=0$ via direct modification of the linear system. For the PINN, we use a fully connected feed-forward neural network (3 hidden layers, 50 neurons per layer, $\tanh$ activation) trained to minimize a composite loss consisting of the Dirichlet boundary residual at $x \in \{-1, 1\}$ (evaluated at the two endpoints) and the PDE residual $|(\Delta+\kappa^2)u - f|^2$ evaluated at 400 interior residual points sampled in the domain $\Omega$. Training uses the Adam optimizer (learning rate $\eta=10^{-3}$) for $10^4$ iterations, with hyperparameters fixed. Additional implementation details of the two baseline solvers are provided in Appendix~\ref{app:baselines}. All errors in Table~\ref{tab:tab1d_real} are computed on a uniform test grid of $N_{\text{test}}=1000$ points over $[-1,1]$.

For context, FDM and PINN use substantially more interior constraints (grid nodes or residual points) than the $N_{\text{col}}\in\{3,5,7\}$ LinPDE-GP settings. We therefore interpret Table~\ref{tab:tab1d_real} primarily as a constraint-efficiency comparison rather than a best-possible accuracy benchmark for LinPDE-GP. In particular, the $N_{\text{col}}=7$ LinPDE-GP solution has lower error than the PINN across all reported metrics and is comparable to the FDM result, while using only seven interior PDE observations.

\begin{table}[t!]
\caption{\label{tab:tab1d_real}1D Real Helmholtz Equation: Comparison of solution accuracy for different numerical and learning-based solvers. Errors are reported as mean squared error (MSE), mean absolute error (MAE), maximum absolute error, and relative $\ell^2$ error. For LinPDE-GP, ($N_{\text{col}}$ points) denotes the number of collocation points.}
\begin{ruledtabular}
\begin{tabular}{lrrrr}
Method & MSE & MAE & Max Error & Rel.\ $\ell^2$ \\
\hline
PINN & 2.73e-05 & 3.90e-03 & 1.26e-02 & 4.26e-03 \\
FDM  & 4.05e-09 & 5.62e-05 & 9.99e-05 & 5.19e-05 \\
LinPDE-GP (3 points) & 6.84e-03 & 7.49e-02 & 1.11e-01 & 6.74e-02 \\
LinPDE-GP (5 points) & 5.83e-06 & 2.29e-03 & 2.77e-03 & 1.97e-03 \\
LinPDE-GP (7 points) & 7.28e-09 & 8.23e-05 & 1.01e-04 & 6.96e-05 \\
\end{tabular}
\end{ruledtabular}
\end{table}

In summary, this 1D real-valued setting shows that conditioning on boundary and PDE observations contracts the posterior in both the operator-applied field and the primal solution, with stronger contraction as the number of PDE collocation points increases. Having validated the method in this controlled environment, we extend this diagnostic view to the complex-valued and higher-dimensional experiments in the subsequent sections.

\subsection{Extension to complex-valued wavenumbers in 1D}

We next consider the one-dimensional case with a complex-valued squared wavenumber, evaluated through the real2 representation that yields a coupled real-valued system for GP conditioning. On $\Omega=[0.0,1.0]$, we solve
$$(\Delta + \kappa^2)\, u(x) = f(x)$$
subject to Dirichlet boundary conditions $u(0) = 1$ and $u(1) = 0$. The wavenumber is complex-valued and arises from a linear viscoelastic constitutive model with $\kappa^2 = \rho\omega^2 / G = 1 - \mathrm{i}$ and having parameters $\rho = 1$, $\omega = 2$, and complex shear modulus $G = 2 + 2\mathrm{i}$. The right-hand side is the complex constant $f(x) = 2 + 3\mathrm{i}$. This configuration admits an analytical solution of the form $u^*(x)=c_1\sin(\kappa x)+c_2\cos(\kappa x)+f/\kappa^2$, where the coefficients $c_1$ and $c_2$ are determined by the boundary data.

Although one-dimensional, this setting is diagnostically informative because $\kappa^2\notin\mathbb{R}$ couples the real and imaginary components through the operator. Writing $u=u_R+\mathrm{i}u_I$ and $\kappa^2=a+\mathrm{i}b$ (here $a=1$, $b=-1$), the PDE is equivalent to the coupled real system
\begin{equation}
\begin{aligned}
    u_R''(x)+a\,u_R(x)-b\,u_I(x)&=\operatorname{Re}(f),\\
    u_I''(x)+a\,u_I(x)+b\,u_R(x)&=\operatorname{Im}(f).
\end{aligned}
\end{equation}
This shows that any posterior dependence between $u_R$ and $u_I$ can be induced by conditioning on the differential equation, even if it is absent a priori.

We employ a zero-mean Gaussian process prior over $u$, mapping the complex-valued unknown to a two-output real function $u(x) \mapsto (\operatorname{Re}(u(x)),\, \operatorname{Im}(u(x)))$. The covariance structure is a scalar exponentiated quadratic (squared exponential) kernel with variance $\sigma_f^2 = 4$ and lengthscale $\ell = 0.7$, lifted to the two-output setting via a shared-kernel construction that assigns $\sigma_f^2/2=2$ to each component, so $u_R$ and $u_I$ each carry marginal prior variance $2$. No component-wise cross-correlations are imposed at the prior level; all coupling between the real and imaginary parts arises exclusively through conditioning on the complex-valued differential operator.

\begin{figure*}[!ht]
    \centering
    \includegraphics[width=0.9\textwidth]{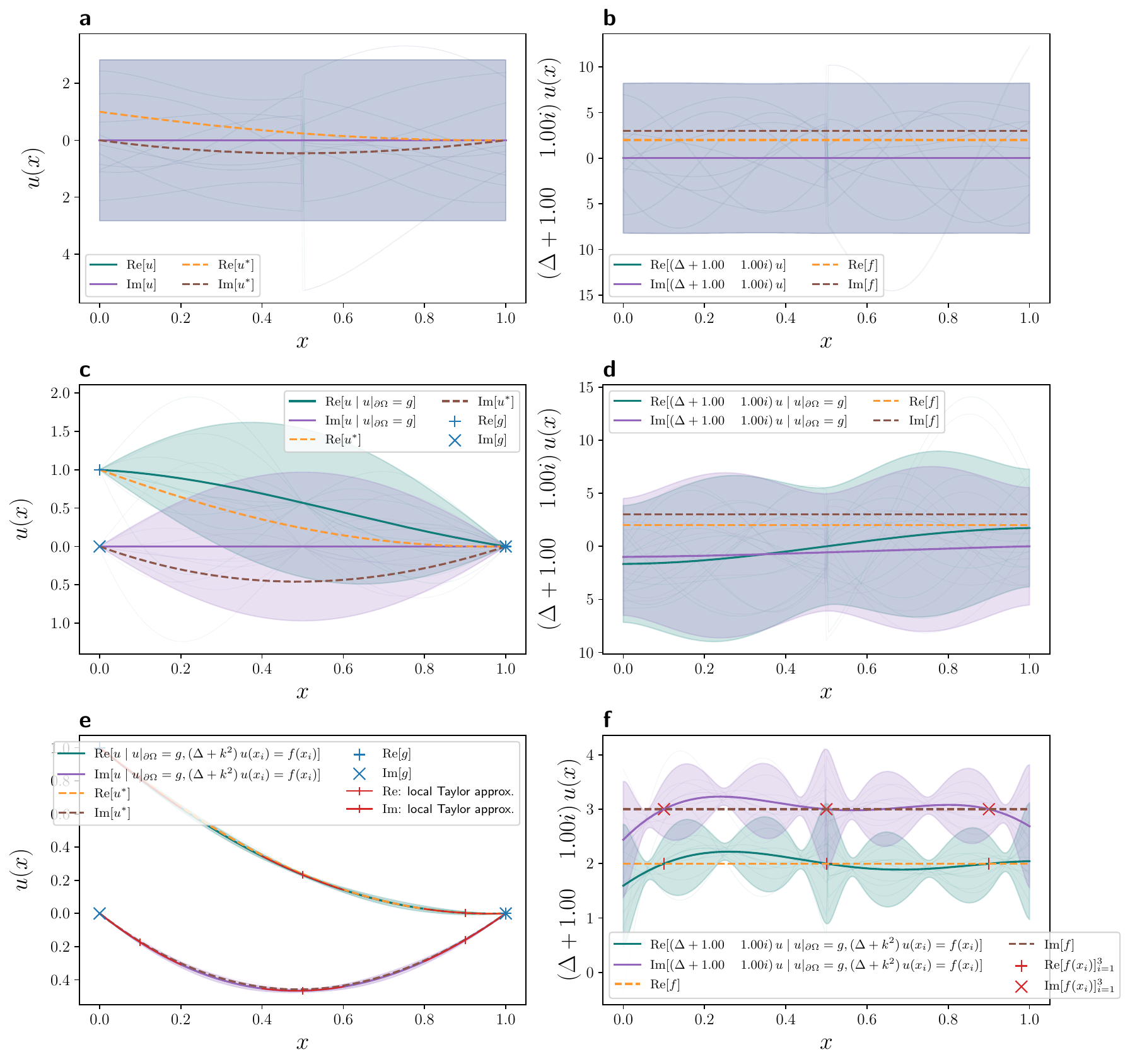}
    \caption{\label{fig:1d_complex_3pts} \textbf{1D Complex Helmholtz Equation: Posterior evolution under sequential conditioning for a complex-valued 1D Helmholtz equation.} Left column: Belief about the solution $u(x)$. Right column: Belief about the operator-applied field $(\Delta + \kappa^{2})u(x)$, overlaid with the source term $f$. Rows: (a,b) prior; (c,d) conditioned on boundary values $g$ at $\partial\Omega$; (e,f) conditioned on boundary values and PDE observations at the collocation points. Blue and purple lines: GP means for $\operatorname{Re}(u)$ and $\operatorname{Im}(u)$ (and their operator-applied counterparts), respectively; Shaded bands: $\pm 2$ posterior standard deviations (approximately 95\% mass under a Gaussian assumption) for each component; faint lines: posterior samples. Dashed lines: analytical solution $u^{*}$ (left) and source term $f$ (right); markers (‘+’ for real, ‘$\times$’ for imaginary) indicate observed components—boundary values $g$ in the left column and collocation values $f(x_i)$ in (f); the short parabolic segments in (e) visualize the local Taylor approximation of the posterior mean at the collocation sites. The squared wavenumber $\kappa^{2}$ is complex-valued, coupling the real and imaginary parts through the Helmholtz operator. Empirical standard deviations of the operator-applied posterior (right column) are estimated from $5\times10^{4}$ Monte Carlo samples and lightly smoothed with a Gaussian filter of standard deviation 2 grid points for visual clarity.}
\end{figure*}

Posterior inference proceeds sequentially as illustrated in Fig.~\ref{fig:1d_complex_3pts}, following the same conditioning protocol as the real-valued experiment to isolate the respective contributions of boundary data and PDE observations. The prior, shown in (a,b), is first conditioned on the two Dirichlet boundary observations, $u(0) = 1 + 0\mathrm{i}$ and $u(1) = 0 + 0\mathrm{i}$, which constrain the function values at the domain endpoints as illustrated in (c,d). The boundary-conditioned process is then further conditioned on the PDE observations at three equispaced interior collocation points (placed on a uniform grid with a boundary inset of 0.1) in (e,f), enforcing $(\Delta + \kappa^2)\,u(x_i) = f(x_i)$ for $i = 1, 2, 3$. This two-stage protocol isolates the respective contributions of the boundary data and the differential-equation constraint, enabling a transparent analysis of how each source of information reshapes the posterior belief.

Conditioning on the Dirichlet data (panels c, d) anchors the posterior at the domain endpoints, collapsing the marginal uncertainty to zero at $x=0$ and $x=1$. Between the boundaries the posterior remains diffuse, with sample paths exhibiting substantial variability. In the operator view (panel d), the induced distribution over $(\Delta + \kappa^{2})u$ remains broadly uncertain and does not yet enforce agreement with the prescribed right-hand side $f = 2 + 3\mathrm{i}$, particularly in the interior. This stage demonstrates that boundary data alone are insufficient to resolve the interior structure of the solution.

Incorporating the differential-equation constraint at three interior collocation points (panels e, f) produces a marked reduction in posterior uncertainty across the entire domain. The posterior mean closely tracks the analytical solution for both the real and imaginary components, and the credible bands contract to a narrow envelope around the ground truth. The posterior samples become nearly indistinguishable from the mean, indicating that the combination of boundary and PDE information is highly informative even with a small number of collocation sites. In the operator view (panel f), the induced distribution over $(\Delta + \kappa^{2})u$ concentrates tightly around the constant right-hand side, with the posterior mean passing through the collocation values and the credible bands tightening around those locations.

In this example, the analytical solution lies within the $\pm 2 \sigma$ credible bands throughout the domain at every conditioning stage, consistent with well-calibrated posterior uncertainty. The progressive narrowing of the bands from prior to full posterior mirrors the information content of each conditioning step and provides a principled quantification of epistemic uncertainty that would be absent in a purely deterministic solver.

\paragraph{Baseline solvers.}
We compare against a FDM and a PINN in Table~\ref{tab:tab1d_complex} under a common evaluation protocol. In both cases, errors are computed on a uniform test grid of $N_{\text{test}}=1000$ points over $[0,1]$ and reported separately for the real part and imaginary part, evaluated against the closed-form analytical solution.

\begin{table*}[!ht]
\caption{\label{tab:tab1d_complex}1D Complex Helmholtz Equation: Comparison of solution accuracy for the 1D complex-valued Helmholtz equation. Errors are reported separately for the real and imaginary components as mean squared error (MSE), mean absolute error (MAE), maximum absolute error, and relative $\ell^2$ error. For LinPDE-GP, $N_{\text{col}}$ denotes the number of interior collocation points.}
\begin{ruledtabular}
\begin{tabular}{llrrrr}
Method & Component & MSE & MAE & Max Error & Rel.\ $\ell^2$ \\
\hline
\multirow{2}{*}{PINN}
  & Real & 7.24e-13 & 6.87e-07 & 1.89e-06 & 1.90e-06 \\
  & Imaginary & 6.57e-12 & 2.13e-06 & 5.06e-06 & 7.68e-06 \\
\hline
\multirow{2}{*}{FDM}
  & Real & 2.45e-15 & 4.51e-08 & 6.83e-08 & 1.11e-07 \\
  & Imaginary & 6.20e-17 & 7.20e-09 & 1.08e-08 & 2.36e-08 \\
\hline
\multirow{2}{*}{LinPDE-GP ($N_{\text{col}}=3$)}
  & Real & 1.84e-05 & 3.19e-03 & 7.70e-03 & 9.59e-03 \\
  & Imaginary & 4.14e-05 & 5.73e-03 & 9.47e-03 & 1.93e-02 \\
\hline
\multirow{2}{*}{LinPDE-GP ($N_{\text{col}}=5$)}
  & Real & 2.35e-10 & 1.24e-05 & 3.43e-05 & 3.43e-05 \\
  & Imaginary & 6.66e-10 & 2.28e-05 & 4.98e-05 & 7.73e-05 \\
\hline
\multirow{2}{*}{LinPDE-GP ($N_{\text{col}}=7$)}
  & Real & 8.00e-13 & 7.73e-07 & 1.67e-06 & 2.00e-06 \\
  & Imaginary & 6.49e-12 & 2.42e-06 & 4.30e-06 & 7.63e-06 \\
\end{tabular}
\end{ruledtabular}
\end{table*}

For FDM, we use a standard second-order central finite-difference discretization on a uniform grid over $[0,1]$ with $N_{\text{fdm}}=1000$ nodes (grid spacing $\Delta x \approx 1.0 \times 10^{-3}$), solving the complex-valued system directly. The interior unknowns satisfy the tridiagonal linear system with main-diagonal entries $-2/\Delta x^2 + \kappa^2$ and off-diagonal entries $1/\Delta x^2$, where $\kappa^2 = \rho\omega^2/(G'+\mathrm{i}G'')$ is the complex-valued squared wavenumber. 
Dirichlet boundary conditions $u(0) = 1$ and $u(1) = 0$ are enforced by direct modification of the right-hand side. The resulting complex-valued linear system is solved by direct factorization. The FDM solution is linearly interpolated onto the test grid for error evaluation. 

For PINN, we use a fully connected feed-forward neural network with architecture $[1, 64, 64, 64, 2]$ (three hidden layers of 64 neurons each, $\tanh$ activation, Glorot normal initialization). Because the solution is complex-valued, the network outputs two channels corresponding to the real and imaginary parts of $u$, and the PDE is decomposed into a coupled real system. Writing $\kappa^2 = a + \mathrm{i}b$ and $f = f_{\mathrm{re}} + \mathrm{i} f_{\mathrm{im}}$, the residuals
$r_{\mathrm{re}} = u_{\mathrm{re}}'' + a\, u_{\mathrm{re}} - b\, u_{\mathrm{im}} - f_{\mathrm{re}}$
and
$r_{\mathrm{im}} = u_{\mathrm{im}}'' + a\, u_{\mathrm{im}} + b\, u_{\mathrm{re}} - f_{\mathrm{im}}$
are jointly minimized. Dirichlet boundary conditions $u(0) = 1$ and $u(1) = 0$ are enforced via point-set constraints at the two endpoints, applied separately to each output component. The composite loss comprises the two PDE residual terms and the two boundary residual terms. The PDE residual is evaluated at $N_{\text{res}} = 998$ fixed interior anchor points placed on a uniform grid over $[0,1]$, with no stochastic resampling. Training is configured in two phases: first, the Adam optimizer ($\eta = 10^{-3}$) for up to $2 \times 10^4$ iterations, then L-BFGS is run to convergence. All hyperparameters are fixed. Additional implementation details are provided in Appendix~\ref{app:baselines}. 

As with the real-valued case, the PINN uses substantially more interior constraints than the $N_{\text{col}} \in \{3, 5, 7\}$ LinPDE-GP settings, and we therefore interpret Table~\ref{tab:tab1d_complex} primarily as a constraint-efficiency comparison. Table~\ref{tab:tab1d_complex} shows that LinPDE-GP accuracy improves systematically as $N_{\text{col}}$ increases, with the $N_{\text{col}}=7$ setting attaining errors comparable to the PINN while using far fewer interior constraints, whereas FDM achieves the smallest errors overall at a much denser discretization.

Having established in one dimension that operator-induced coupling between real and imaginary components can be captured from only a few PDE observations, we next turn to a two-dimensional complex-valued problem. This setting provides a more stringent test of whether the same sparse-conditioning strategy remains accurate when the solution geometry, spatial correlations, and operator constraints become higher-dimensional and more structured.

\subsection{Complex-valued Helmholtz problem in two dimensions}
\label{subsec:2d_complex}

We now consider the complex-valued Helmholtz problem on the square domain $\Omega=[-1,1]^2$ with homogeneous Dirichlet boundary conditions. As in the one-dimensional complex case, the squared wavenumber arises from the linear viscoelastic constitutive model, $\kappa^2 = \rho\omega^2/G$, here with $\rho=1$, $\omega=1$, and the complex shear modulus $G = 1 + 0.5\mathrm{i}$, yielding $\kappa^2 = 0.8 - 0.4\mathrm{i}$ (i.e., $a=0.8$, $b=-0.4$). The forcing is the purely real constant $f(\mathbf{x}) = 2$. Writing $u = u_R + \mathrm{i}u_I$, the coupled real system introduced in the one-dimensional setting carries over verbatim, with the second derivative replaced by the Laplacian. Because $b \neq 0$, the operator couples $u_R$ and $u_I$, so even a purely real forcing must induce a nonzero imaginary component in the solution.

The prior follows the one-dimensional complex construction with a two-dimensional base kernel. The base kernel is a tensor product of exponentiated quadratic (squared-exponential) kernels, with length scale $\ell = 1$ per dimension and output variance $\sigma_f^2 = 4$. It is lifted to a block-diagonal two-output kernel over $(u_R, u_I)$, with each channel assigned $\sigma_f^2/2 = 2$. Thus $u_R$ and $u_I$ each carry marginal prior variance $2$, and the prior imposes no cross-correlation between the components. Any posterior dependence between them arises exclusively through conditioning on the complex-valued operator. Inference then follows the collocation protocol of the real-valued experiment. Homogeneous Dirichlet data are imposed through $N_{\mathrm{bc}} = 20$ boundary observations per edge ($80$ in total), placed on $\partial\Omega$ with their tangential coordinate inset from the corners by $10^{-6}$. Each boundary site constrains both the real and imaginary components of $u$. The PDE is enforced through operator observations $(\Delta + \kappa^2)u(\mathbf{x}_i) = f(\mathbf{x}_i)$ on a uniform $12 \times 12$ interior collocation grid ($N_{\mathrm{col}} = 144$), inset from $\partial\Omega$ by $10^{-6}$ in both dimensions. Each collocation site contributes one real and one imaginary constraint. All observations carry independent Gaussian jitter, with variance $10^{-7}$ for boundary observations and $10^{-4}$ for operator observations. The larger operator jitter reflects the poorer conditioning of the second-order operator Gram matrix.

\begin{figure*}[!htbp]
    \centering
    \includegraphics[width=0.6\textwidth]{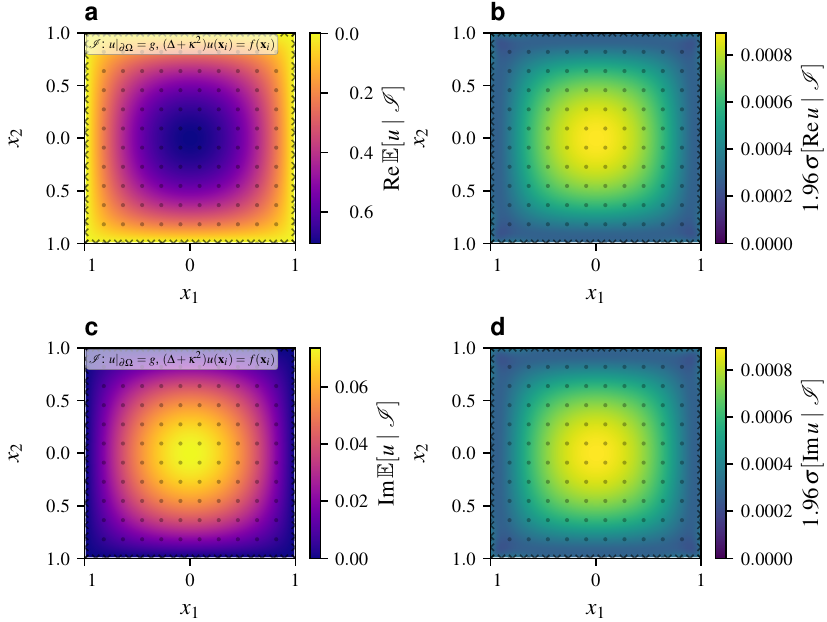}
    \caption{\label{fig:2d_complex_posterior} \textbf{Posterior solution and uncertainty for the 2D complex-valued problem.} Rows correspond to the real part $u_R$ (top) and imaginary part $u_I$ (bottom) of the solution. Left column (a,c): posterior mean fields $\operatorname{Re}\,\mathbb{E}[u(\mathbf{x}) \mid \mathcal{I}]$ and $\operatorname{Im}\,\mathbb{E}[u(\mathbf{x}) \mid \mathcal{I}]$ evaluated on a dense visualization grid over $\Omega$. Right column (b,d): pointwise $95\%$ credible half-width $1.96\,\sigma[\cdot \mid \mathcal{I}]$ for each component on the same grid (i.e., $1.96$ times the posterior standard deviation, not the full interval width). Boundary-condition locations (crosses) and PDE collocation locations (dots) are overlaid to indicate where the posterior has been conditioned and to visually relate uncertainty reduction to the imposed information set $\mathcal{I}$. The nonzero imaginary mean in (c) is generated entirely by the complex squared wavenumber $\kappa^2 = 0.8 - 0.4\mathrm{i}$, since both the forcing and the boundary data are purely real and the prior places no cross-correlation between the components.}
\end{figure*}

Fig.~\ref{fig:2d_complex_posterior} summarizes the posterior over both components. The posterior mean of the real part (panel a) retains the smooth, symmetric bowl of the real-valued experiment, vanishing on $\partial\Omega$ and reaching its most negative value near the center of the domain. The imaginary part (panel c) shares the same radial symmetry but has opposite curvature. Its mean forms a positive dome that peaks near the center of the domain and is roughly an order of magnitude smaller. This component is entirely operator-induced: with a purely real forcing and a block-diagonal prior, only the coupling term $b\,u_R$ in the differential equation can generate it. The credible half-widths (panels b and d) are largest in the central portion of the domain and decay toward the boundary for both components. This reflects the dense Dirichlet conditioning that pins both components directly on $\partial\Omega$. Interior field values are constrained only indirectly, since operator observations probe derivatives of the field rather than field values directly.

\begin{figure*}[!htbp]
    \centering
    \includegraphics[width=0.85\textwidth]{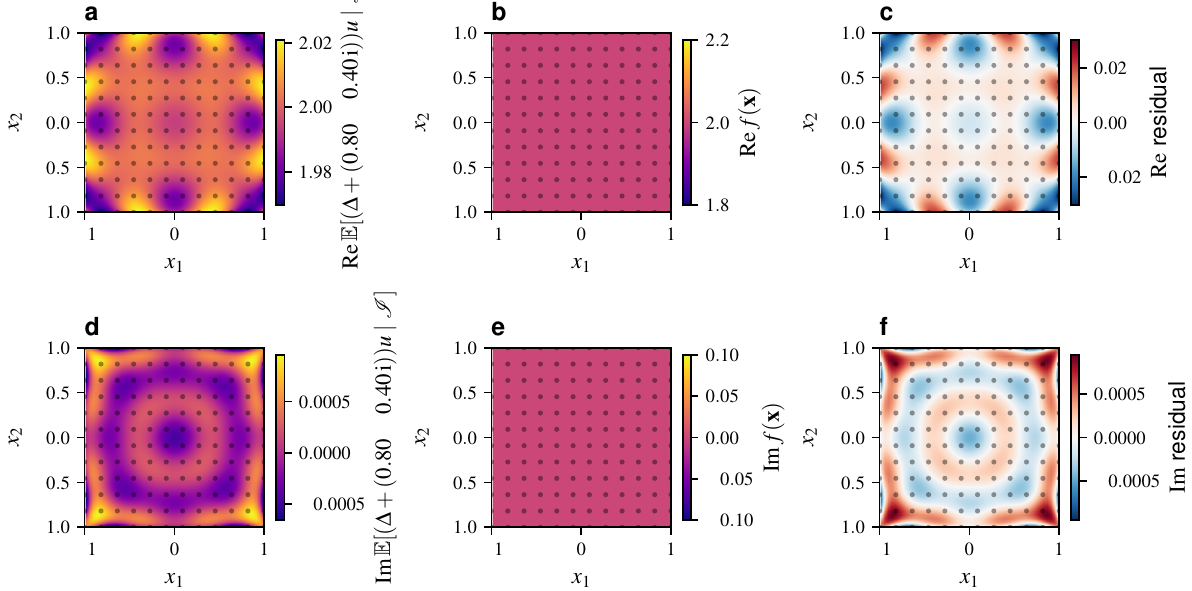}
    \caption{\label{fig:2d_complex_residual} \textbf{Posterior mean PDE satisfaction diagnostics for the coupled system.} Rows correspond to the real (top) and imaginary (bottom) components of the complex Helmholtz equation. Each row assesses how closely the posterior mean satisfies its equation by differencing two fields on a dense visualization grid: (a,d) the operator-applied posterior mean $\mathbb{E}[(\Delta + \kappa^2)u(\mathbf{x}) \mid \mathcal{I}]$, obtained by applying the coupled Helmholtz operator to the posterior GP and taking its mean; (b,e) the corresponding component of the forcing term $f(\mathbf{x}) = 2 + 0\mathrm{i}$; and (c,f) the mean residual, defined as the difference of the two preceding columns. Only the third column (c,f) is the residual itself; the first two columns show the operator-applied mean and the forcing that are subtracted to form it. Overlaid PDE collocation points mark where the operator constraints were imposed. The residual reports how closely the posterior mean satisfies both equations of the coupled system across $\Omega$, not the uncertainty of the residual.}
\end{figure*}

Fig.~\ref{fig:2d_complex_residual} assesses PDE satisfaction of posterior means, separately for the two components of the coupled system. Across most of $\Omega$, both components of the operator-applied posterior mean closely match the corresponding components of the forcing. The residuals remain small but spatially structured. The real residual (panel c) oscillates between collocation points across the interior. The imaginary residual (panel f) instead concentrates its largest deviations at the corners and boundary-adjacent regions. These residual structures differ from the posterior uncertainty, which peaks in the interior. The residual instead measures how well the posterior mean satisfies the operator, which is enforced only at the discrete collocation points. It is therefore largest in the gaps between those points and near the corners, rather than where the marginal variance is greatest. Notably, the imaginary residual (panel f) remains small even though the imaginary equation must balance the coupling term $b\,u_R$ against $\Delta u_I + a\,u_I$ to reproduce a vanishing imaginary forcing.

The posterior mean was further validated against the analytical solution. For a constant complex source with homogeneous Dirichlet data, this solution admits the sine eigenfunction expansion of Eq.~\eqref{eqn:nd_analytic} with $d=2$, carrying the complex $\kappa^2$ through $(-\lambda_{mn} + \kappa^2)^{-1}$. The expansion is truncated to odd modes up to index $99$ per direction (fifty modes per axis). On a $31 \times 31$ evaluation grid, the posterior mean attains relative $\ell^2$ errors of $1.72 \times 10^{-3}$ (real) and $1.68 \times 10^{-4}$ (imaginary), with maximum pointwise errors of $2.90 \times 10^{-3}$ (real) and $1.86 \times 10^{-5}$ (imaginary). The coupled formulation therefore recovers the operator-induced imaginary component to high accuracy despite its small amplitude. This confirms that conditioning on the complex Helmholtz operator drives the posterior mean toward domain-wide satisfaction of both equations of the coupled system.

We next compare the posterior mean against deterministic baseline solvers.

\paragraph{Baseline solvers.} We compare against an FDM and a PINN in Table~\ref{tab:tab2d_complex}, using the same $31 \times 31$ evaluation grid and analytical reference as above. Errors are reported separately for the real and imaginary parts. For FDM, we use a standard second-order five-point discretization of $\Delta + \kappa^2$ on a uniform grid with $N_{\text{fdm}} = 161$ nodes per dimension, assembled as a single complex-valued sparse linear system so that the complex $\kappa^2$ and source enter natively.

For PINN, we use a fully connected feed-forward neural network with architecture $[2, 64, 64, 64, 2]$ (three hidden layers of 64 neurons each, $\tanh$ activation, Glorot normal initialization). As in the one-dimensional case, the network outputs two channels corresponding to the real and imaginary parts of $u$, and the PDE is decomposed into a coupled real system. Writing $\kappa^2 = a + \mathrm{i}b$ and $f = f_{\mathrm{re}} + \mathrm{i}f_{\mathrm{im}}$, the residuals $r_{\mathrm{re}} = \Delta u_{\mathrm{re}} + a\,u_{\mathrm{re}} - b\,u_{\mathrm{im}} - f_{\mathrm{re}}$ and $r_{\mathrm{im}} = \Delta u_{\mathrm{im}} + a\,u_{\mathrm{im}} + b\,u_{\mathrm{re}} - f_{\mathrm{im}}$ are jointly minimized. In contrast to the one-dimensional case, the homogeneous Dirichlet boundary conditions are enforced as hard constraints via the output transform $\hat{u}(x_1, x_2) = (1-x_1^2)(1-x_2^2)\,\mathrm{NN}_{\theta}(x_1, x_2)$, applied to both output channels, so the loss comprises the two PDE residual terms alone, evaluated at $N_{\text{res}} = 2000$ interior collocation points. Training follows the same two-phase schedule: the Adam optimizer ($\eta = 10^{-3}$) for $2 \times 10^4$ iterations, then L-BFGS run to its stopping criterion. All hyperparameters are fixed. Additional implementation details are provided in Appendix~\ref{app:baselines}.

\begin{table*}[!ht]
\caption{\label{tab:tab2d_complex}2D Complex Helmholtz Equation: Comparison of solution accuracy for the 2D complex-valued Helmholtz equation. Errors are reported separately for the real and imaginary components as mean squared error (MSE), mean absolute error (MAE), maximum absolute error, and relative $\ell^2$ error. For LinPDE-GP, $N_{\text{col}}$ denotes the number of interior collocation points.}
\begin{ruledtabular}
\begin{tabular}{llrrrr}
Method & Component & MSE & MAE & Max Error & Rel.\ $\ell^2$ \\
\hline
\multirow{2}{*}{PINN}
  & Real & 8.61e-12 & 1.98e-06 & 2.91e-05 & 7.78e-06 \\
  & Imaginary & 4.76e-12 & 1.58e-06 & 7.84e-06 & 6.00e-05 \\
\hline
\multirow{2}{*}{FDM}
  & Real & 1.10e-10 & 9.14e-06 & 1.86e-05 & 2.78e-05 \\
  & Imaginary & 1.96e-13 & 3.87e-07 & 6.20e-07 & 1.22e-05 \\
\hline
\multirow{2}{*}{LinPDE-GP ($N_{\text{col}}=8^2$)}
  & Real & 3.50e-07 & 3.72e-04 & 2.71e-03 & 1.57e-03 \\
  & Imaginary & 1.69e-10 & 1.20e-05 & 2.08e-05 & 3.57e-04 \\
\hline
\multirow{2}{*}{LinPDE-GP ($N_{\text{col}}=10^2$)}
  & Real & 3.90e-07 & 3.40e-04 & 2.84e-03 & 1.66e-03 \\
  & Imaginary & 3.69e-11 & 4.54e-06 & 1.72e-05 & 1.67e-04 \\
\hline
\multirow{2}{*}{LinPDE-GP ($N_{\text{col}}=12^2$)}
  & Real & 4.20e-07 & 3.40e-04 & 2.90e-03 & 1.72e-03 \\
  & Imaginary & 3.73e-11 & 4.24e-06 & 1.86e-05 & 1.68e-04 \\
\end{tabular}
\end{ruledtabular}
\end{table*}

As shown in Table~\ref{tab:tab2d_complex}, in this two-dimensional complex-valued setting, the deterministic baselines remain highly accurate, with the PINN and FDM reaching relative $\ell^2$ errors of order $10^{-6}$ to $10^{-5}$ in both components. LinPDE-GP attains a relative $\ell^2$ error of $1.57$ to $1.72\times 10^{-3}$ in the real part. Its imaginary-part error is smaller, between $1.67$ and $3.57\times 10^{-4}$. This is about $4\times$ below the real-part error at $N_{\text{col}}=8^2$ and roughly an order of magnitude below the errors corresponding to $N_{\text{col}}=10^2$ and $12^2$. The gap partly reflects the smaller amplitude of the operator-induced imaginary component, against which the relative error is normalized. The real-part error does not improve as $N_{\text{col}}$ grows from $8^2$ to $12^2$, increasing slightly from $1.57\times 10^{-3}$ to $1.72\times 10^{-3}$. This insensitivity to the collocation budget indicates that accuracy in this range is limited more by the prior and model specification than by the number of collocation points. Unlike the point estimates of the baselines, the GP posterior also supplies marginal uncertainty over both components, a property exploited next in the three-dimensional complex-valued experiment.

\subsection{Complex-valued Helmholtz problem in three dimensions}

We complete the synthetic study with the complex-valued Helmholtz problem on the cube $\Omega=[-1,1]^3$ with homogeneous Dirichlet boundary conditions, which tests whether sparse operator conditioning scales to a full volumetric domain. As in the lower-dimensional complex cases, the squared wavenumber arises from a linear viscoelastic constitutive model, $\kappa^2=\rho\omega^2/G$, here with $\rho=1$, $\omega=1$, and the complex shear modulus $G=1+0.5\mathrm{i}$, yielding $\kappa^2=0.8-0.4\mathrm{i}$ (i.e., $a=0.8$, $b=-0.4$). The forcing is the purely real constant $f(\mathbf{x})=2$. Writing $u=u_R+\mathrm{i}u_I$, the coupled real system introduced in the one-dimensional setting carries over verbatim, with the Laplacian now acting in three dimensions. Because $b\neq0$, the operator couples $u_R$ and $u_I$, so the real forcing again induces a nonzero imaginary component in the solution.

The prior follows the one-dimensional complex construction with a three-dimensional base kernel. The base kernel is a tensor product of exponentiated quadratic (squared-exponential) kernels, with length scale $\ell=1$ per dimension and output variance $\sigma_f^2=4$. It is lifted to a block-diagonal two-output kernel over $(u_R, u_I)$, with each channel assigned $\sigma_f^2/2=2$. Thus $u_R$ and $u_I$ each carry marginal prior variance $2$, and the prior imposes no cross-correlation between the components. Any posterior dependence between them arises exclusively through conditioning on the complex-valued operator. Inference then follows the collocation protocol of the lower-dimensional experiments. Homogeneous Dirichlet data are imposed on a $12\times12$ lattice on each of the six cube faces ($N_{\mathrm{bc}}=144$ per face, $864$ in total), placed on $\partial\Omega$ with an inset of $10^{-4}$ in the face coordinates. Each boundary site constrains both the real and imaginary components of $u$. The PDE is enforced through operator observations $(\Delta+\kappa^2)u(\mathbf{x}_i)=f(\mathbf{x}_i)$ on a regular $10\times10\times10$ interior collocation grid ($N_{\mathrm{col}}=1000$), inset from $\partial\Omega$ by $10^{-6}$ in all three dimensions. Each collocation site contributes one real and one imaginary constraint. All observations carry independent Gaussian jitter, with variance $10^{-6}$ for boundary observations and $10^{-4}$ for operator observations. The larger operator jitter reflects the poorer conditioning of the second-order operator Gram matrix.

\begin{figure}[!htbp]
    \centering
    \includegraphics[width=\columnwidth]{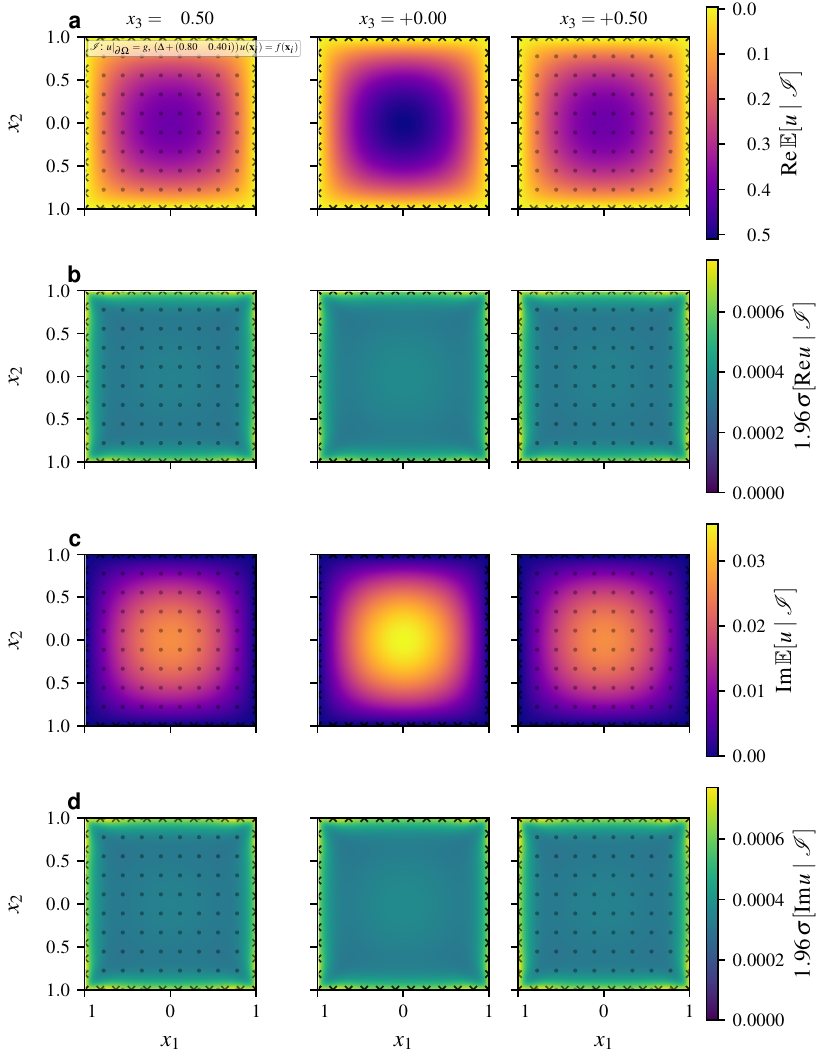}
    \caption{\label{fig:3d_complex_posterior} \textbf{Multi-plane posterior slices of the 3D complex solution field.} Columns are axial slices $x_3=z$. Rows show, from top to bottom, the posterior mean of the real part (\textbf{a}) and its $95\%$ credible half-width $1.96\,\sigma$ (\textbf{b}), then the posterior mean of the imaginary part (\textbf{c}) and its credible half-width (\textbf{d}), all conditioned on the information set $\mathcal{I}$ ($864$ boundary and $1000$ interior PDE observations), summarized by the in-panel annotation in the top-left panel. Conditioning locations are overlaid on each slice (PDE collocation points as semi-transparent black dots and boundary-condition points as black crosses, selected by proximity in $x_3$), with the domain outline drawn in white. The nonzero imaginary mean is generated entirely by the complex squared wavenumber $\kappa^2=0.8-0.4\mathrm{i}$, since both the forcing and the boundary data are purely real and the prior places no cross-correlation between the components.}
\end{figure}

Fig.~\ref{fig:3d_complex_posterior} visualizes the inferred field through planar posterior slices at $x_3\in\{-0.5,0,0.5\}$, with the real part in the upper two rows and the imaginary part below. For both components the posterior mean is smooth and respects the homogeneous boundary condition, approaching zero toward the cube faces and attaining its largest magnitude near the center of the domain. The real mean recovers the familiar negative bowl of the lower-dimensional experiments. The imaginary mean, generated entirely by the coupling term $b\,u_R$, forms a positive dome of the same symmetry at roughly an order of magnitude smaller amplitude. The accompanying uncertainty maps show that the pointwise $95\%$ credible half-width is uniformly small throughout the bulk and rises near faces, edges, and corners. This boundary-localized uncertainty is consistent with the discrete sampling of Dirichlet data on each face and the placement of interior operator observations on a grid inset from $\partial\Omega$. Both leave the regions closest to the boundary comparatively less constrained.

\paragraph{Baseline solvers.}
We compare against an FDM and a PINN under a common evaluation protocol. All errors are computed on a shared $21\times21\times21$ uniform grid over $\Omega=[-1,1]^3$ ($9{,}261$ points) and reported separately for the real and imaginary parts, evaluated against the analytical solution of Eq.~\eqref{eqn:nd_analytic} with $d=3$ (odd modes up to index $49$ per direction, twenty-five modes per axis), in which $\kappa^2$ enters through the modal denominators $(-\lambda_{mnp}+\kappa^2)^{-1}$. For FDM, we use a standard second-order seven-point discretization of $\Delta+\kappa^2$ assembled as a single complex-valued sparse linear system, so the complex $\kappa^2$ and source enter natively, with homogeneous Dirichlet conditions enforced by direct modification of the boundary rows. Because sparse direct factorization becomes prohibitive in three dimensions, we obtain the FDM solution on a fine $161^3$ grid using a discrete-sine-transform separable solver that diagonalizes the same seven-point operator in $O(N\log N)$, then interpolate it onto the shared evaluation grid; this is the three-dimensional analogue of the one- and two-dimensional complex FDM references.

For PINN, we use a fully connected feed-forward network with architecture $[3,64,64,64,2]$ (three hidden layers of $64$ neurons each, $\tanh$ activation, Glorot normal initialization). As in the lower-dimensional cases, the two output channels represent the real and imaginary parts and the complex PDE is split into the coupled real system, with residuals $r_{\mathrm{re}}=\Delta u_{\mathrm{re}}+a\,u_{\mathrm{re}}-b\,u_{\mathrm{im}}-f_{\mathrm{re}}$ and $r_{\mathrm{im}}=\Delta u_{\mathrm{im}}+a\,u_{\mathrm{im}}+b\,u_{\mathrm{re}}-f_{\mathrm{im}}$ jointly minimized. Homogeneous Dirichlet conditions are imposed as hard constraints through the output transform $\hat{u}(\mathbf{x})=(1-x_1^2)(1-x_2^2)(1-x_3^2)\,\mathrm{NN}_{\theta}(\mathbf{x})$ applied to both channels, so the loss comprises the two PDE residual terms alone, evaluated at $N_{\text{res}}=8000$ interior collocation points. Training follows the two-phase schedule used throughout: the Adam optimizer ($\eta=10^{-3}$) for $2\times10^4$ iterations, then L-BFGS run to its stopping criterion. All hyperparameters are fixed. Additional implementation details are provided in Appendix~\ref{app:baselines}.

As in two dimensions, the deterministic baselines remain highly accurate (Table~\ref{tab:tab3d_complex}), with PINN and FDM relative $\ell^2$ errors of order $10^{-5}$ to $10^{-4}$ in both components. LinPDE-GP attains a relative $\ell^2$ error of approximately $7.6\times10^{-3}$ in the real part. The imaginary part is recovered six- to sevenfold more accurately, at $1.1$--$1.2\times10^{-3}$. The real-part error is essentially unchanged as $N_{\text{col}}$ grows from $9^3$ to $11^3$, again indicating that accuracy in this regime is governed by the prior and model specification rather than by the collocation budget. Although the deterministic solvers remain more accurate, by roughly two orders of magnitude in the real part, LinPDE-GP conditions on at most $11^3$ interior constraints, compared with $8{,}000$ collocation points for the PINN and a $161^3$ reference grid for the FDM. Furthermore, unlike either baseline, the posterior additionally supplies marginal uncertainty over both components.

\begin{table*}[!ht]
\caption{\label{tab:tab3d_complex}3D Complex Helmholtz Equation: Comparison of solution accuracy for the 3D complex-valued Helmholtz equation. Errors are reported separately for the real and imaginary components as mean squared error (MSE), mean absolute error (MAE), maximum absolute error, and relative $\ell^2$ error. For LinPDE-GP, $N_{\text{col}}$ denotes the number of interior collocation points.}
\begin{ruledtabular}
\begin{tabular}{llrrrr}
Method & Component & MSE & MAE & Max Error & Rel.\ $\ell^2$ \\
\hline
\multirow{2}{*}{PINN}
  & Real      & $1.04\times10^{-10}$ & $6.65\times10^{-6}$ & $2.14\times10^{-4}$ & $4.92\times10^{-5}$ \\
  & Imaginary & $1.91\times10^{-11}$ & $2.93\times10^{-6}$ & $1.68\times10^{-5}$ & $3.53\times10^{-4}$ \\
\hline
\multirow{2}{*}{FDM}
  & Real      & $2.65\times10^{-10}$ & $1.20\times10^{-5}$ & $5.38\times10^{-5}$ & $7.86\times10^{-5}$ \\
  & Imaginary & $9.85\times10^{-14}$ & $1.76\times10^{-7}$ & $1.19\times10^{-6}$ & $2.53\times10^{-5}$ \\
\hline
\multirow{2}{*}{LinPDE-GP ($N_{\text{col}}=9^3$)}
  & Real      & $2.48\times10^{-6}$  & $9.93\times10^{-4}$ & $6.17\times10^{-3}$ & $7.61\times10^{-3}$ \\
  & Imaginary & $1.89\times10^{-10}$ & $1.03\times10^{-5}$ & $4.17\times10^{-5}$ & $1.11\times10^{-3}$ \\
\hline
\multirow{2}{*}{LinPDE-GP ($N_{\text{col}}=10^3$)}
  & Real      & $2.48\times10^{-6}$  & $9.88\times10^{-4}$ & $6.17\times10^{-3}$ & $7.60\times10^{-3}$ \\
  & Imaginary & $2.11\times10^{-10}$ & $1.08\times10^{-5}$ & $4.39\times10^{-5}$ & $1.17\times10^{-3}$ \\
\hline
\multirow{2}{*}{LinPDE-GP ($N_{\text{col}}=11^3$)}
  & Real      & $2.48\times10^{-6}$  & $9.85\times10^{-4}$ & $6.17\times10^{-3}$ & $7.60\times10^{-3}$ \\
  & Imaginary & $2.30\times10^{-10}$ & $1.12\times10^{-5}$ & $4.55\times10^{-5}$ & $1.22\times10^{-3}$ \\
\end{tabular}
\end{ruledtabular}
\end{table*}

Taken together, the synthetic experiments trace a consistent picture across spatial dimensions and across the transition from real to complex wavenumbers. In one dimension, conditioning on a handful of boundary and PDE observations contracts the posterior in both the operator-applied field and the solution, and the complex case shows that the realified solver recovers both components of the coupled system from the same sparse information. The two- and three-dimensional complex problems confirm that this sparse-conditioning strategy persists as the geometry, spatial correlations, and operator structure grow richer. There, with purely real forcing and a block-diagonal prior, the imaginary component is generated by the operator coupling alone, and it is recovered to high relative accuracy. The posterior-mean point estimates remain competitive with dense finite-difference and neural-network baselines while using far fewer interior constraints. Throughout, the posterior also reports the approximation uncertainty that the deterministic baselines cannot, and on the one-dimensional complex problem the analytic solution lies within the $\pm2\sigma$ credible bands at every conditioning stage, consistent with well-calibrated uncertainty in that setting.

The benchmarking problems considered so far establish the performance and accuracy of the proposed method in controlled settings where ground truth and exact coefficients are available. We now apply it to a setting where neither holds: an \textit{in vivo} human-brain MRE reconstruction with a spatially varying, independently estimated $\kappa^2(\mathbf{x})$. This reconstruction tests the method under measurement noise, highly complex geometry, physical model mismatch, and vector-field curl observables. Here the model mismatch is unavoidable, since the measured field only approximately satisfies the idealized scalar Helmholtz model.

\section{Three-dimensional complex-valued Helmholtz reconstruction on in vivo brain MRE data}
\label{exp:brain}

We apply the complex Helmholtz solver to \textit{in vivo} human brain MRE data from the publicly available Brain Biomechanics Imaging Resources (BBIR) repository in the NeuroImaging Tools and Resources Collaboratory (NITRC) \citep{bayly2021mr}; Appendix~\ref{app:dataset} details the full provenance and preprocessing. The present study uses three subjects (0001, 0002, and 0003). For each subject the BBIR repository provides a complex-valued, time-harmonic displacement field measured by MRE at $1.5$\,mm isotropic spatial resolution and at mechanical drive frequencies of $30$, $50$, and $70$\,Hz, together with nonlinear-inversion (NLI) maps of the storage and loss shear moduli $G'(\mathbf{x})$ and $G''(\mathbf{x})$. We do not condition on the raw displacement. Instead, we use the curl $\mathbf{q}=\nabla\times\mathbf{u}$ supplied with the release, which removes the irrotational (compressional) part and any rigid-body motion. The curl is a vector field, and each of its three Cartesian components satisfies a scalar Helmholtz equation of the same form as Eq.~\eqref{eqn:helmholtz} under the local-homogeneity assumption of Sec.~\ref{sec:formu_n_notation}. We reconstruct one component at a time, so ``curl component'' below denotes which Cartesian component of $\mathbf{q}$ is reconstructed. The components are ordered $x$, $y$, $z$, with $y$ axis corresponding to the anterior--posterior direction.

From the NLI moduli we form the spatially varying squared wavenumber $\kappa^2(\mathbf{x}) = \rho\,\omega^2/\!\left(G'(\mathbf{x}) + \mathrm{i}\,G''(\mathbf{x})\right)$, with $\rho = 1040\,\mathrm{kg\,m^{-3}}$ and $\omega = 2\pi f_{\mathrm{drive}}$~\citep{upadhyay2022development, upadhyay2022data}. This fixes the variable-coefficient operator $\Delta + \kappa^2(\mathbf{x})$ at each drive frequency. For a given subject, frequency, and component, we place a Mat\'ern-$5/2$ prior in the proper real2 construction of Sec.~\ref{subsec:real2}. We then condition on two information sets: Dirichlet observations of the boundary trace, sampled on a surface shell of the brain mask ($N_{\mathrm{shell}}=2000$), and interior collocation sites that enforce the homogeneous residual $(\Delta + \kappa^2(\mathbf{x}))\,q = 0$ ($N_{\mathrm{int}}=2000$). No interior field value enters as an observation, so the reconstructed interior field is an out-of-sample test of physical consistency rather than a fit to held-in data. The measured field has no closed-form solution and only approximately obeys the idealized scalar model. We therefore quantify the agreement between the predicted and measured curl magnitude by their Pearson correlation, with an acceptance target of $\mathrm{Pearson}>0.75$.

We track model mismatch separately through the empirical Helmholtz residual, computed from the measured curl and the fixed coefficient $\kappa^2(\mathbf{x})$, independent of the GP reconstruction. Applying the assumed operator to the curl gives the local residual $r(\mathbf{x}) = (\Delta + \kappa^2(\mathbf{x}))\,q(\mathbf{x})$, which vanishes wherever $q$ satisfies the homogeneous scalar Helmholtz relation exactly. Normalizing its magnitude by $|\kappa^2 q|$ yields the dimensionless diagnostic $|r|/|\kappa^2 q|$, reported as the median over interior voxels. Both magnitudes are taken over the three Cartesian curl components, so the diagnostic is a single value shared by the components. The ratio is non-negative and has no upper bound, in contrast to the Pearson correlation, which is confined to $[-1,1]$. Because $\kappa^2 q$ is one of the two terms the homogeneous relation balances, unity is its natural reference scale. A value near $0$ indicates a well-specified model, with the residual negligible against $|\kappa^2 q|$. A value of $1$ means the residual is as large as $|\kappa^2 q|$ itself. Values above $1$, as found throughout the brain data (median $\approx 3$ to $4$ at $70$\,Hz), indicate that the residual exceeds $|\kappa^2 q|$ and the scalar homogeneous model is strongly violated.

Throughout this section the prior is \emph{multiscale}: a two-scale linear model of coregionalization (LMC) that combines two base length scales. Its base kernel sums a fine ($4$\,mm) and a coarse ($16$\,mm) Mat\'ern-$5/2$ component, with zero real--imaginary correlation ($\rho_{\mathrm{RI}}=0$; Sec.~\ref{subsec:real2}). We call a reconstruction obtained with this prior the \emph{multiscale reconstruction}. We run it on the full grid of three subjects, three drive frequencies, and three curl components ($3\times3\times3=27$ configurations). Rather than tabulate all twenty-seven cells, we report the results by varying \emph{one-factor-at-a-time} (OFAT) about a fixed \emph{reference} configuration (subject 0001, $70$\,Hz, component $x$). From this reference we vary a single factor while holding the other two fixed, sweeping subject, then drive frequency, then curl component. This isolates each factor's main effect on agreement; interaction effects across the grid are not analyzed (see Limitations). A subsequent ablation then varies the prior itself to identify which part of the multiscale construction drives the gain. We read each generalization trend against the model-mismatch diagnostic $|r|/|\kappa^2 q|$, and finally assess whether the posterior uncertainty is calibrated.

\begin{figure*}[t]
    \centering
    \includegraphics[width=0.91\textwidth]{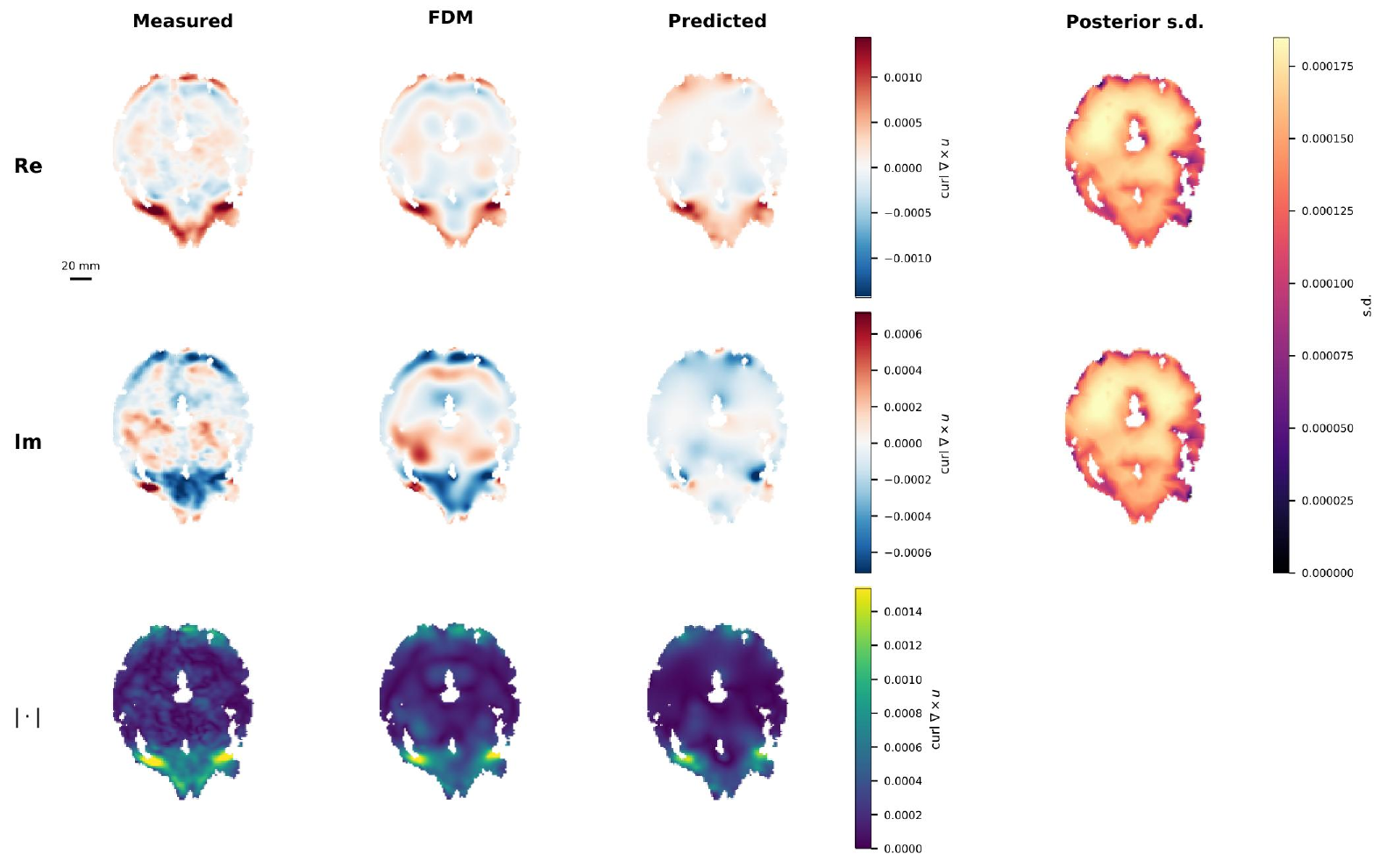}
    \caption{\label{fig:brain_recon} \textbf{GP--PDE reconstruction of the brain curl wavefield.}
    Gaussian-process solution of the variable-coefficient Helmholtz forward problem for subject 0003 at 70~Hz. The target is the $x$-component of the curl observable $q=\nabla\times u$, reconstructed from 2000 boundary-shell and 2000 interior collocation points. The prior is a Mat\'ern ($\nu=2.5$) two-scale linear model of coregionalization, with base length scales 4 and 16~mm and zero Re/Im correlation ($\rho_{\mathrm{RI}}=0$). All panels show the same central axial slice; the brain mask removes background and interior non-brain (e.g.\ ventricular) voxels. Rows give the real part (Re), imaginary part (Im) and magnitude ($|\cdot|$) of the time-harmonic field; columns compare the measured field, a deterministic finite difference method (FDM) solution of the same Helmholtz problem, and the GP posterior mean (Predicted). The FDM column fixes the same $\kappa^2(\mathbf{x})$ and imposes the measured curl as the Dirichlet boundary trace, with the interior held out. Within each row the three panels share one color scale, fixed to the 99th percentile of the measured field, so the posterior-mean amplitude shrinkage is shown rather than hidden by autoscaling. Re and Im use a diverging scale (red positive, blue negative) and $|\cdot|$ a sequential (viridis) scale. The right column maps the posterior standard deviation (Posterior s.d.) for the Re and Im channels on a shared magma scale. The magnitude row has no s.d.\ panel, because $|q|$ is a nonlinear function of the two Gaussian-process outputs and the solver does not produce its variance. Scale bar: 20~mm. The reconstruction reaches a Pearson correlation of $0.753$ between predicted and measured curl magnitude (median relative error $0.400$), clearing the $0.75$ acceptance target. The deterministic FDM solution, which imposes the full measured boundary, reaches a higher agreement ($0.847$; median relative error $0.299$). It too stays well short of unity, so the residual gap reflects the scalar model rather than the inference. The empirical Helmholtz residual has median $|r|/|\kappa^2 q| = 2.98$, and the posterior standard deviation tracks this residual only weakly ($\mathrm{corr}(\sigma,r)=0.33$).}
\end{figure*}

Table~\ref{tab:brain_generalization} reports the multiscale LMC reconstruction across the three axes of variation, each varied one factor at a time starting with the reference configuration (subject 0001, $70$\,Hz, component $x$). At the reference, the reconstruction attains a Pearson correlation of $0.767$ between the predicted and measured curl magnitude (median relative error $0.375$), clearing the acceptance target of $0.75$. Similarly, for subject 0003 at $70$\,Hz, component $x$, the reconstruction clears the acceptance target (Pearson $0.753$). Fig.~\ref{fig:brain_recon} compares the posterior-mean solution for this case with the corresponding measured curl field and with a deterministic finite-difference solve of the same Helmholtz model.

Across the three subjects at $70$\,Hz and component $x$, the method generalizes but not uniformly. Subjects 0001 and 0003 clear the target ($0.767$ and $0.753$), whereas subject 0002 falls short at $0.679$. Over the same subjects, the empirical Helmholtz residual is broadly comparable ($|r|/|\kappa^2 q| = 3.65$, $3.70$, and $2.98$). The subject-to-subject variation in Pearson correlation therefore does not track model mismatch, and instead reflects subject-level differences in the measured curl field.

\begin{table}[!ht]
  \centering
  \caption{Generalization of the multiscale LMC curl reconstruction across subjects, drive frequencies, and curl components (one factor at a time about the reference configuration; prior LMC with base length scales $4/16$\,mm and Re/Im correlation $\rho_{\mathrm{RI}}=0$). Pearson is $\mathrm{corr}(|\hat q|,|q|)$ on the $n=2000$ collocation set; ``rel.\ err'' is the median $\big||\hat q|-|q|\big|/|q|$ (relative error of the magnitude); and the residual diagnostic is the median $|r|/|\kappa^2 q|$ of the empirical Helmholtz residual. The acceptance target is $\mathrm{Pearson}>0.75$. Boldface marks the reference configuration (subject 0001, $70$\,Hz, component $x$) about which the sweep is centered.}
  \label{tab:brain_generalization}
  \begin{tabular}{c c c r r r}
    \toprule
    Subject & Freq.\ (Hz) & Comp. & Pearson & rel.\ err & $|r|/|\kappa^2 q|$ \\
    \midrule
    0001 & 70 & $x$ & \textbf{0.767} & 0.375 & 3.65 \\
    \addlinespace[2pt]
    0002 & 70 & $x$ & 0.679 & 0.387 & 3.70 \\
    0003 & 70 & $x$ & 0.753 & 0.400 & 2.98 \\
    \addlinespace[2pt]
    0001 & 30 & $x$ & 0.596 & 0.346 & 6.07 \\
    0001 & 50 & $x$ & 0.735 & 0.346 & 3.75 \\
    \addlinespace[2pt]
    0001 & 70 & $y$ & 0.591 & 0.533 & 3.65 \\
    0001 & 70 & $z$ & 0.626 & 0.432 & 3.65 \\
    \bottomrule
  \end{tabular}
\end{table}

The frequency sweep exposes a physical ceiling that is not removed by tuning. Holding subject 0001 and component $x$ fixed, the agreement rises with drive frequency, from $0.596$ at 30\,Hz to $0.735$ at 50\,Hz and $0.767$ at 70\,Hz. The 30\,Hz case reconstructs worst and also carries the largest empirical Helmholtz residual ($|r|/|\kappa^2 q| = 6.07$, against $3.75$ at 50\,Hz and $3.65$ at 70\,Hz). Because this residual is computed directly from the measured field, independently of the GP reconstruction, the larger value at 30\,Hz shows that the scalar local-homogeneity model fits the data least well at low frequency, a mismatch the solver cannot overcome \citep{THOMASSEALE20161781, perreard2010effects}. The 50\,Hz reconstruction sits just below the target at $0.735$, identifying 50\,Hz as a moderate ceiling and 70\,Hz as the regime in which the model is best specified.

This frequency trend has a physical origin in the shear wavelength. The local-homogeneity reduction keeps the term $\kappa^2 q$ and neglects the modulus-gradient coupling, of order $(\nabla G/G)\cdot\nabla q$. Relative to the retained term, this neglected term scales as $(|\nabla G|/G)\,\lambda$, the fractional change in modulus over one shear wavelength $\lambda = 2\pi/\mathrm{Re}(\kappa)$. The wavelength shortens as drive frequency rises, so this fractional change shrinks and the local-homogeneity approximation tightens. At $30$\,Hz the wavelength is long and approaches the scale over which the tissue modulus varies. The neglected term is then a large fraction of the wave operator, and the empirical residual is correspondingly higher ($|r|/|\kappa^2 q| = 6.07$). At $70$\,Hz the wavelength is short, the medium looks locally homogeneous over a wavelength, and the residual is smallest. The monotone improvement over $30$ to $70$\,Hz follows this scaling, although it need not continue to higher frequencies, where attenuation and reduced displacement amplitude lower the curl signal-to-noise ratio.

The component sweep separates the shared model mismatch from per-component data quality. At subject 0001 and $70$\,Hz, we evaluate the empirical residual on the full curl vector field, which yields a single value shared by the three Cartesian components ($|r|/|\kappa^2 q| = 3.65$). The scalar operator $\Delta+\kappa^2$ acts component-wise, so a per-component residual $|r_i|/(|\kappa^2|\,|q_i|)$ is also well defined; we report the vector form because the per-component residual mixes model mismatch with the lower signal-to-noise ratio of the noisier curl components. The shared vector residual, therefore, does not, on its own, resolve the spread in agreement across components ($0.767$, $0.626$, and $0.591$ for $x$, $z$, and $y$). We attribute that spread to differences in the measured curl data between components, with the anterior--posterior component $y$ lowest. The median relative error stays within a narrow band across the cases in Table~\ref{tab:brain_generalization} ($0.346$ to $0.533$), confirming that the cases differ little in amplitude accuracy. The wide spread in agreement therefore reflects how much of the field's spatial structure each reconstruction recovers, the quantity the Pearson correlation measures.

Table~\ref{tab:brain_prior_ablation} isolates the role of the prior at the reference configuration and shows that the multiscale base kernel, not Re/Im coupling, drives the gain. A single-scale independent (IID) prior plateaus at a Pearson of $0.686$, below the acceptance target. Replacing it with a two-scale LMC at fine and coarse length scales of $4$ and $16$\,mm raises the agreement to $0.767$ and clears the target, while the $3/16$\,mm pairing reaches $0.747$, just short of the target. Introducing a non-zero Re/Im correlation does not help: setting $\rho_{\mathrm{RI}} = 0.6$ yields $0.761$, no improvement over the proper $\rho_{\mathrm{RI}} = 0$ prior at $0.767$. The proper, multiscale prior ($\rho_{\mathrm{RI}} = 0$, Sec.~\ref{subsec:real2}) therefore both clears the target and stays within the proper (circular) family, whereas the cross-channel coupling that leaves that family confers no benefit.

\begin{table}[t]
  \centering
  \caption{Prior ablation at the reference configuration (subject 0001, 70\,Hz, component $x$). A single-scale independent (IID) prior plateaus below the acceptance target; a two-scale linear model of coregionalization (LMC) with fine/coarse length scales $4/16$\,mm clears it. Adding a non-zero Re/Im correlation $\rho_{\mathrm{RI}}$ does not improve point accuracy, so the gain originates from the multiscale kernel rather than from Re/Im coupling. The last column reports calibration as $\mathrm{corr}$ of the posterior standard deviation with the Helmholtz residual $|r|/|\kappa^2 q|$.}
  \label{tab:brain_prior_ablation}
  \begin{tabular}{l c c r r r}
    \toprule
    Prior & Length scales (mm) & $\rho_{\mathrm{RI}}$ & Pearson & rel.\ err & $\mathrm{corr}(\sigma, r)$ \\
    \midrule
    IID & $15$ (single) & --  & 0.686 & 0.448 & 0.474 \\
    LMC & $3,\,16$      & 0   & 0.747 & 0.389 & 0.364 \\
    LMC & $4,\,16$      & 0   & \textbf{0.767} & 0.375 & 0.347 \\
    LMC & $4,\,16$      & 0.6 & 0.761 & 0.379 & 0.352 \\
    \bottomrule
  \end{tabular}
\end{table}

The point-accuracy gain does not, however, come with improved uncertainty calibration. The diagnostic correlation between the posterior standard deviation and the local Helmholtz residual is modest throughout, and it weakens as accuracy improves, falling from $0.474$ for the IID prior to $0.347$ for the champion LMC. Adding the Re/Im coupling leaves it essentially unchanged ($0.352$ at $\rho_{\mathrm{RI}} = 0.6$). The posterior standard deviation thus flags where the physics is violated only weakly, and the same coupling that fails to improve point accuracy also fails to repair calibration. The diagnostic correlation is, moreover, identical across the three components at fixed subject and frequency. The full sweep returns $0.347$ for $x$, $y$ and $z$ at subject 0001, 70\,Hz, of which Table~\ref{tab:brain_prior_ablation} lists the $x$ entry. This follows because the GP posterior variance is set by the conditioning geometry, kernel, and operator, not by the observed values. The reported uncertainty therefore cannot separate the well-reconstructed component ($x$, Pearson $0.767$) from the poorly reconstructed one ($y$, Pearson $0.591$). We read the posterior width as a qualitative indicator rather than a calibrated error bar, and return to this point in the Discussion.

Taken together, the brain reconstruction succeeds where the scalar curl model is well specified, clearing the acceptance target at the reference configuration and for two of the three subjects at 70\,Hz. Its boundaries are set not by the solver but by the physics and by uncertainty calibration. The local-homogeneity model fits the data least well at low drive frequency, capping the agreement there, and the posterior standard deviation tracks model mismatch only weakly, a deficit that Re/Im coupling does not remove. We take up both boundaries and their bearing on whether the posterior can be trusted in the Discussion.

\section{Limitations}

The brain reconstruction rests on a scalar local-homogeneity model anchored by measured boundary data. We treat each Cartesian component of the curl field as an independent solution of the same scalar Helmholtz equation. The squared wavenumber $\kappa^2(\mathbf{x})$ is fixed from independently inverted modulus maps, and Dirichlet data are imposed on a surface shell of the brain mask. Five assumptions follow, three on the interior model and two on the boundary. 1. The medium is locally homogeneous, so spatial gradients of the modulus are neglected. 2. The compressional motion is fully removed by the curl, leaving a purely shear field. 3. The modulus maps are exact, so their inversion error is not propagated. 4. The mask surface acts as a true Dirichlet boundary, although it truncates a larger domain rather than physically bounding the wavefield. 5. The boundary trace and the mask geometry are both treated as exact, so neither measurement noise nor segmentation error is propagated. The two boundary idealizations are especially consequential on real tissue. \textit{In vivo}, skull-driven motion is transmitted to the parenchymal brain tissue through a complex boundary involving the meninges and cerebrospinal fluid (CSF), yet the present masked-domain Helmholtz reconstruction does not impose explicit transmission, absorbing, or radiation conditions at this interface. The measured trace, itself noisy, enters with only a small numerical jitter rather than a calibrated noise model. The shell that carries it is a subsampled, binary-eroded approximation of the segmentation, and the MRE data are least reliable near the skull-brain interface \citep{murphy2013measuring}. Only the interior assumptions can be checked against the measured field. The measured curl satisfies the homogeneous relation $|r|/|\kappa^2 q|<1$ at fewer than about $6\%$ of interior voxels at $70$\,Hz, with a median ratio of roughly $3$ to $4$. Assumptions 1--3 therefore hold only weakly, and the boundary idealizations remain untested, an additional source of model error. The solver can reproduce the dominant spatial structure of the field but not its fine detail. The achievable agreement is bounded primarily by the model rather than by the inference.
 
Inference uses exact Gaussian-process conditioning, whose cost grows cubically with the number of observations. Each brain solve conditions on a subsampled set of $2000$ boundary and $2000$ interior sites rather than the full voxel grid. Because every site carries a real and an imaginary channel, this amounts to $8000$ scalar observations assembled into a dense two-channel system. On the CPU used here, where the GPU kernel back end was unavailable, a single component takes roughly $25$ to $45$\,s to solve and a further $15$ to $25$\,s to evaluate the posterior mean and a subsampled posterior standard deviation on a dense slice; evaluating the standard deviation exactly over the same slice is far costlier, of order $300$\,s. This scaling limits both the spatial resolution we can condition on and the size of the generalization study. The solver is also numerically delicate in how the observation noise is scaled. The prior output scale $\sigma_f$ is tuned to the field magnitude, of order $10^{-4}$ for the curl (so the prior variance $\sigma_f^2$ is of order $10^{-8}$) against $O(1)$ in the benchmarks. The observation noise enters the Gram diagonal at that scale, so it is meaningful only as a fraction of it. We therefore set the boundary and PDE noises as fractions of model-derived variances, $10^{-6}$ of the prior variance and $10^{-4}$ of the conditioned prior variance respectively. A fixed absolute floor would instead dominate the constraint contribution when $\sigma_f^2$ is small and leave the Gram ill-conditioned when it is large.

The evidence base is modest. We study one publicly available dataset, three subjects, three drive frequencies, and three curl components. We compute the full subject $\times$ frequency $\times$ component grid, but report it as one-factor-at-a-time variations about a single anchor, and we do not analyze interaction effects. The synthetic benchmarks that carry the baseline comparison are constant-coefficient problems with known reference solutions, and they include complex-valued cases in one, two, and three dimensions. The brain experiment is the only variable-coefficient problem and the only one on measured data. It therefore has no independent ground-truth field, because the measured curl does not solve the idealized model and no exact reference exists for the tissue. A deterministic finite-difference solve of the same model, with $\kappa^2(\mathbf{x})$ fixed and the measured trace as Dirichlet data, supplies a known solution of that model. We show it as the FDM column of Fig.~\ref{fig:brain_recon}. For that subject the deterministic solve reaches a higher agreement than the GP (Pearson $0.847$ against $0.753$). It nonetheless stays well short of unity, which locates the dominant ceiling in the scalar local-homogeneity model rather than in the inference. The GP trades this margin for two properties the deterministic solve lacks. It conditions on a sparse, scattered set of $2000$ boundary and $2000$ interior sites drawn from the measured MRE voxel grid, whereas the deterministic solve resolves that same grid in full ($307{,}518$ interior unknowns and an $88{,}650$-voxel Dirichlet shell for the subject shown), and it returns a posterior over the field. We therefore keep the GP as the primary solver and read the deterministic solution as a model-error reference. A full separation of inference error from model error remains future work.
 
The posterior uncertainty does not function as a calibrated error bar on real data. Its correlation with the local model residual is modest at best and weakens as point accuracy improves, from $0.47$ for the single-scale prior to $0.35$ for the multiscale champion, and it falls to $0.05$ at $30$\,Hz. Because the Gaussian-process posterior variance is fixed by the conditioning geometry, the kernel, and the operator, and not by the observed values, it is identical across the three curl components even though their agreement ranges from $0.59$ to $0.77$. Filtering predictions by posterior confidence does not isolate the accurate ones. The high-confidence subset can carry larger relative error than the full set, and for the multiscale champion it numbers only about $30$ of the $2000$ evaluation sites. Finally, the solver returns variances for the real and imaginary channels but not for the curl magnitude that the agreement metric uses, because the magnitude is a nonlinear function of the two Gaussian outputs. The reported uncertainty should therefore be read as a qualitative map of where information is sparse, not as a trustworthy error estimate.

\section{Discussion}

This work extends operator-informed Gaussian-process inference from real-valued to complex-valued and dissipative wave problems, and it carries that machinery from synthetic benchmarks onto measured data. The probabilistic-numerics view of linear PDE solvers, formalized for real fields by \citet{pförtner2024physicsinformedgaussianprocessregression}, treats the solution as Bayesian inference and returns a posterior rather than a point estimate. By realifying the complex Helmholtz operator and placing a proper or coregionalized prior on the real and imaginary channels, we retain that posterior for fields whose attenuation requires a complex modulus. This connects two literatures that have developed apart. PDE-oriented physics-informed GP regression has largely been real-valued, while complex-valued GP regression matured in signal processing without an operator-conditioning view.
 
The brain ablation gives a concrete design lesson for complex GP-PDE priors. Point accuracy is driven by the multiscale structure of the base kernel, not by a priori coupling of the real and imaginary channels. Adding a fine $4$\,mm scale to a coarse $16$\,mm scale lifts the agreement above the acceptance target, whereas introducing an improper cross-channel correlation leaves it essentially unchanged. We interpret this to mean that the curl wavefield carries information at more than one spatial scale, which a single length scale cannot represent, while the real--imaginary dependence the data demand is already induced by the dissipative operator during conditioning. A proper, multiscale prior is therefore both sufficient and preferable, since it clears the target without leaving the circular family.
 
That the reconstruction succeeds at all is informative, because the measured field departs substantially from the homogeneous scalar model it is conditioned on. Conditioning on a residual the data only weakly satisfy still recovers the dominant spatial pattern. This suggests that the boundary trace and the low-order operator structure constrain the field more than the exact local balance does. The point-accuracy finding is robust. It holds for two of three subjects at $70$\,Hz and strengthens monotonically with drive frequency. The frequency trend is physical rather than numerical, because the empirical residual, computed from the data alone, is largest at $30$\,Hz, where the scalar local-homogeneity model fits worst. The solver cannot overcome a model that the data violate, and the low-frequency ceiling marks that boundary.
 
The uncertainty results carry a cautionary message for probabilistic PDE solvers on real data. A posterior is useful for downstream decisions only if its spread reflects actual error, and here it does not. The variance is set by where we placed observations, not by how well the physics holds, so it cannot flag the regions or components where the reconstruction is least trustworthy. The improper prior we expected might repair this did not, which locates the problem in the noise and likelihood model rather than in the prior covariance. We therefore read the present posterior as a structured interpolation-uncertainty map, and we caution against using it as a calibrated confidence bound, especially in a biomedical setting where overconfidence carries real cost.
 
Several directions follow from these boundaries. Calibration is the most pressing. A noise model that scales with the local Helmholtz residual, or hyperparameters learned by marginal likelihood, could let the posterior reflect model mismatch rather than sampling geometry. The independent treatment of the three curl components could be replaced by a jointly coupled vector model. The missing magnitude variance could be recovered by propagating the two-channel posteriors through the nonlinear magnitude, by sampling or a delta-method approximation~\citep{oehlert1992note}. On real tissue, the dominant source of mismatch is the homogeneity assumption. A tissue-specific parameterization that distinguishes gray and white matter, with spatially varying length scales and modulus-gradient terms, would address it directly. Finally, scalable Gaussian-process techniques would lift the resolution and sample-size limits that exact conditioning imposes. Inducing points, iterative solvers, or GPU kernel evaluation could also enable joint inference of the modulus field rather than treating it as fixed.

\section{Conclusion}

We have extended operator-informed Gaussian-process regression to complex-valued Helmholtz problems through a real--imaginary realification, equipped it with a family of proper and coregionalized priors, and applied it from analytic benchmarks through to measured \textit{in vivo} brain elastography. On benchmark problems in one to three dimensions with known reference solutions, the solver matches the finite-difference and neural-network baselines in one dimension and trails them in higher dimensions. It does so while conditioning on far fewer interior constraints, and, unlike those baselines, it returns a posterior over the complex wavefield rather than a single estimate. On real brain data a proper multiscale prior reconstructs the shear curl field above the acceptance target, and the gain comes from the multiscale base kernel rather than from real--imaginary coupling. The same experiments mark the method's boundaries: a low-frequency ceiling set by the physics, and a posterior uncertainty that does not yet function as a calibrated error bar. Bringing physics-informed probabilistic inference to complex, dissipative media is, we believe, an important step for viscoelastic material characterization, and turning its uncertainty into a trustworthy one is the natural next target.

\begin{acknowledgments}
This material is based upon work supported by the National Science Foundation under Grant Nos. 2331295 and 2623467. The authors gratefully acknowledge the insights and data shared by Prof. Curtis Johnson and Olivia Bailey from the University of Delaware.
\end{acknowledgments}

\section*{Code Availability Statement}
The code used to generate the main results and figures of this study is publicly available at [https://github.com/bydeng01/linpde-gp], and is released under the MIT License. This implementation builds upon the original \textbf{LinPDE-GP} software (available at [https://github.com/marvinpfoertner/linpde-gp], which is also distributed under the MIT License. Appropriate credit, licensing information, and citations to the original work are provided both in this manuscript and in the associated GitHub repository.

\section*{Data Availability Statement}
The data that support the findings of this study are openly available in the Brain Biomechanics Imaging Resources (BBIR) repository on the NeuroImaging Tools and Resources Collaboratory (NITRC) at
\url{https://www.nitrc.org/projects/bbir}. These data were collected at the University of Delaware and are not original to this study; the acquisition is described by \citet{bayly2021mr}.

\appendix

\section{Notation}
\label{app:notation}
\small
\renewcommand{\arraystretch}{1.05}
\tablefirsthead{%
\toprule
\textbf{Notation} & \textbf{Description} \\
\midrule}
\tablehead{%
\toprule
\textbf{Notation} & \textbf{Description} \\
\midrule}
\tabletail{\bottomrule}
\tablelasttail{\bottomrule}
\topcaption{\label{tab:notation}Summary of notation used in this work.}
\newcolumntype{M}{>{\(}l<{\)}}
\begin{supertabular}{@{}M p{0.68\columnwidth}@{}}
\multicolumn{2}{@{}l}{\textit{Gaussian process regression (real-valued)}} \\[2pt]
\mathcal{X} & Input space \\
x, x' & Input locations in \(\mathcal{X}\) \\
h(x) & Latent (real-valued) function \\
m(x) & GP mean function \\
k(x,x') & Covariance function (kernel) \\
\mathcal{GP}(m,k) & GP with mean \(m\) and kernel \(k\) \\
N & Number of training observations \\
\mathbf{x}_i,\, y_i & Training input and observed target \\
X,\, X_* & Training and test input matrices \\
\mathbf{y} & Vector of observed targets \\
\mathbf{h}_* & Latent function values at test inputs \\
K(X,X') & Covariance (Gram) matrix \\
\bar{\mathbf{h}}_* & Posterior predictive mean vector \\
\mathbb{V}[\mathbf{h}_*] & Posterior predictive covariance \\
\varepsilon_i & Observation noise, \(\varepsilon_i\sim\mathcal{N}(0,\sigma_\varepsilon^2)\) \\
\sigma_\varepsilon^2 & Noise variance \\
I_N & \(N\times N\) identity matrix \\
\alpha & Dual (representer) coefficients \\
\midrule
\multicolumn{2}{@{}l}{\textit{Complex-valued GP regression}} \\[2pt]
h_R(x),\, h_I(x) & Real and imaginary parts of \(h(x)\) \\
\mathrm{i} & Imaginary unit \\
c(x,x') & Pseudo-covariance function \\
(\cdot)^H & Hermitian (conjugate) transpose \\
\mathcal{CN}(\mathbf{m},K) & Proper complex Gaussian distribution \\
\midrule
\multicolumn{2}{@{}l}{\textit{LinPDE-GP framework}} \\[2pt]
u & Unknown solution field\\
\Omega & Spatial domain \\
\partial\Omega & Boundary of \(\Omega\) \\
\mathbb{B} & Banach space containing \(u\) \\
\mathbb{U} & Solution (function) space \\
\mathcal{L} & Bounded linear information operator, \(\mathcal{L}\colon\mathbb{B}\to\mathbb{R}^n\) \\
n & Observation-space dimension \\
\varepsilon & Noise vector, \(\varepsilon\sim\mathcal{N}(\mu,\Sigma)\) \\
\mu,\,\Sigma & Noise mean and covariance \\
(\cdot)^+ & Moore--Penrose pseudoinverse \\
m^{u|y},\, k^{u|y} & Posterior GP mean and covariance \\
\mathcal{D}^{(w)} & PDE operator (strong or weak form) \\
f^{(w)} & Right-hand side function or functional \\
l & Test functional in the dual space \\
\hat{\mathbb{U}} & Trial (finite-dimensional) subspace \\
\mathcal{P}_{\hat{\mathbb{U}}} & Projection onto \(\hat{\mathbb{U}}\) \\
\phi^{(j)} & Trial basis functions, \(j=1,\dots,M\) \\
M & Number of trial basis functions \\
\mathcal{R}_{l,\mathcal{P}_{\hat{\mathbb{U}}}} & MWR information operator \\
\midrule
\multicolumn{2}{@{}l}{\textit{Real2 representation}} \\[2pt]
u_R(x),\, u_I(x) & Real and imaginary parts of \(u(x)\) \\
\tilde{u} & Real2 field, \(\tilde{u}=(u_R,u_I)^\top\) \\
\Phi & Isomorphism \(u\mapsto\tilde{u}\) \\
\tilde{m}(x) & Real2 GP mean, \(\mathbb{R}^2\)-valued \\
\tilde{k}(x,x') & Real2 matrix-valued kernel, \(\mathbb{R}^{2\times 2}\)-valued \\
k_{\mathrm{RR}},\,k_{\mathrm{RI}},\,k_{\mathrm{IR}},\,k_{\mathrm{II}} & Cross-covariance blocks of \(\tilde{k}\) \\
\tilde{\mathcal{L}} & \(2\times 2\) block operator on \(\tilde{u}\) \\
\mathcal{L}_{ij} & Real-valued blocks of \(\tilde{\mathcal{L}}\) \\
B,\, B_q & Coregionalization matrix (PSD); per-component in LMC \\
\rho_{\mathrm{RI}} & Real--imaginary channel correlation \\
Q & Number of LMC components \\
k_q,\, \ell_q,\, \sigma_q^2 & Per-component base kernel, length scale, variance \\
\mathcal{I} & Conditioning information set (as in \(\mathbb{E}[u\mid\mathcal{I}]\)) \\
\midrule
\multicolumn{2}{@{}l}{\textit{Helmholtz instantiation}} \\[2pt]
\mathcal{H} & Helmholtz operator, \(\Delta+\kappa^2\) \\
\Delta & Spatial Laplacian \\
\kappa^2 & Squared wavenumber (\(\in\mathbb{C}\) in general) \\
a,\, b & \(\operatorname{Re}(\kappa^2)\) and \(\operatorname{Im}(\kappa^2)\) \\
\tilde{\mathcal{H}} & Real2 Helmholtz block operator \\
I_2 & \(2\times 2\) identity matrix \\
f(x) & Source/forcing term (default \(f\equiv 0\)) \\
g & Dirichlet boundary data (\(u=g\) on \(\partial\Omega\)) \\
c_1,\, c_2 & Coefficients in the complex-exponential closed-form solution \\
\kappa^2(\mathbf{x}) & Spatially varying squared wavenumber \\
\rho & Mass density \\
\omega & Angular drive frequency, \(\omega=2\pi f_{\mathrm{drive}}\) \\
f_{\mathrm{drive}} & Mechanical drive frequency (Hz) \\
G',\, G'' & Storage and loss shear moduli \\
G & Complex shear modulus, \(G'+\mathrm{i}G''\) \\
\mathbf{u},\, \mathbf{q} & Displacement field and curl observable \(\mathbf{q}=\nabla\times\mathbf{u}\) \\
q & Cartesian component of \(\mathbf{q}\) (scalar curl field per axis) \\
\hat{q} & Posterior-mean (predicted) scalar curl component \\
r & Local Helmholtz residual (diagnostic \(|r|/|\kappa^2 q|\)) \\
\midrule
\multicolumn{2}{@{}l}{\textit{Kernels and hyperparameters}} \\[2pt]
\ell & Kernel length scale \\
\sigma_f,\, \sigma_f^2 & Kernel amplitude / output variance \\
\nu & Mat\'ern smoothness parameter \\
\sigma_{\mathrm{obs}}^2 & Observation (jitter) variance \\
\sigma & Posterior standard deviation (in \(\mathrm{corr}(\sigma,r)\)) \\
\midrule
\multicolumn{2}{@{}l}{\textit{Evaluation, discretization, and sample counts}} \\[2pt]
N_{\text{col}} & Interior PDE collocation points \\
N_{\text{bc}} & Boundary observations \\
N_{\text{test}} & Test-grid evaluation points \\
u^* & Closed-form / reference (ground-truth) solution \\
N_{\mathrm{shell}},\, N_{\mathrm{int}} & Brain boundary-shell and interior collocation counts \\
\midrule
\multicolumn{2}{@{}l}{\textit{Baseline solvers}} \\[2pt]
N_{\text{fdm}} & Finite-difference node count \\
\Delta x\,(\Delta y,\Delta z) & Finite-difference grid spacing \\
N_{\text{res}} & PINN interior residual points \\
\mathrm{NN}_{\theta} & PINN network with weights \(\theta\) \\
\eta & Adam learning rate \\
\end{supertabular}

\section{Description of the in vivo brain MRE dataset}
\label{app:dataset}
 
\subsection{Source, provenance, and acknowledgment}
\label{app:dataset:source}
 
The data analyzed in this study is part of the publicly distributed \emph{Brain Biomechanics Imaging Resources} (BBIR) collection on the NeuroImaging Tools and Resources Collaboratory (NITRC); the subset used here is the \textit{in vivo} human-brain MRE data acquired at the University of Delaware (UDEL) \citep{bayly2021mr}. The resource is maintained by Johns Hopkins University and Washington University, and its collection was supported by the U.S.\ National Institutes of Health (NIH NINDS U01 NS112120 and R56 NS055951). We gratefully acknowledge the University of Delaware investigators who collected these measurements and the BBIR maintainers who curate and openly distribute the resource. The acquisition protocol, the raw-data processing, and the nonlinear-inversion (NLI) reconstruction of the material properties are documented with the release. The NITRC resource is listed under the GNU General Public License v3; the authors claim no ownership of the data, use it for academic research in accordance with that license, and do not redistribute it as part of this work. All scientific credit for the measurements and for the derived property maps belongs to the original investigators and their institutions; the present work contributes only the wavefield-reconstruction methodology applied to the data.
 
\subsection{Role of the dataset within this study}
\label{app:dataset:use}
 
We use the dataset to demonstrate operator-conditioned Gaussian-process reconstruction of a measured shear wavefield, and not to reproduce the nonlinear inversion (NLI) that recovers the modulus maps. The two problems are distinct, and we treat the modulus maps as a fixed, given coefficient field. From the NLI moduli we form the spatially varying squared wavenumber of the shear-wave Helmholtz equation,
\begin{equation}
    \kappa^2(\mathbf{x}) = \frac{\rho\,\omega^2}{G'(\mathbf{x}) + \mathrm{i}\,G''(\mathbf{x})},
    \label{eq:dataset_wavenumber}
\end{equation}
with mass density $\rho = 1040\,\mathrm{kg\,m^{-3}}$ and angular drive frequency $\omega = 2\pi f_{\mathrm{drive}}$.
 
Rather than conditioning on the raw displacement, we condition on the measured curl field $\mathbf{q}(\mathbf{x}) = \nabla\times\mathbf{u}(\mathbf{x})$, also provided in the repository. The curl removes the irrotational compressional component and any rigid-body motion, leaving Cartesian components that each satisfy the same scalar Helmholtz equation under the local-homogeneity assumption. For each subject, frequency, and Cartesian component, we place a Mat\'ern-$5/2$ Gaussian-process prior in the proper real-2 construction, and we condition on two information sets: Dirichlet observations of the boundary trace sampled on the mask surface ($N_{\mathrm{shell}} = 2000$), and interior collocation sites enforcing the homogeneous residual $(\Delta + \kappa^2(\mathbf{x}))\,q = 0$ ($N_{\mathrm{int}} = 2000$). No interior field measurement enters as an observation, so the recovered interior field is an out-of-sample test of physical consistency rather than a fit to held-in data.
 
\subsection{Preprocessing, adaptation, and modifications}
\label{app:dataset:preproc}
 
We made no changes to the source NIfTI files; all adaptation occurs downstream in our analysis pipeline. The brain mask is taken from the anatomy-to-MRE registration directory, in which the segmentation has already been rigidly registered and resampled to the MRE grid ($1.5$\,mm isotropic voxels, $160\times160\times80$ array) by the data providers, giving a one-to-one correspondence between mask and MRE voxels. The reconstruction domain is the masked brain interior embedded in its bounding box. Boundary trace sites are obtained by binary erosion of the mask to isolate a surface shell, and both the boundary and interior site sets are subsampled to the target counts given above. When evaluating the finite-difference Laplacian, the $1.5$\,mm voxel spacing is converted to meters so that $\kappa^2$ (Eq.~\eqref{eq:dataset_wavenumber}) and the differential operator share consistent SI units. The complex displacement and curl fields are handled through the real2 realification of the complex operator, and each Cartesian component is processed independently.

\section{Baseline solver configurations}\label{app:baselines}
 
\subsection{Finite-difference method (FDM)}\label{app:fdm}
The four synthetic benchmarks of Sec.~\ref{sec:exp_n_results} share a single finite-difference baseline, configured per experiment as summarized in Table~\ref{tab:fdm_settings}. In every case we discretize the Helmholtz operator $\mathcal{H}=\Delta+\kappa^{2}$ with the standard second-order central stencil on a uniform grid and impose the Dirichlet data by direct modification of the assembled linear system. The resulting systems are solved by direct factorization or, in three dimensions, by an algebraically equivalent fast-diagonalization method. The scheme is deterministic and returns a point estimate, which provides the mesh-based reference for the posterior mean of LinPDE-GP.
\paragraph{Discretization.}
On a one-dimensional grid with spacing $\Delta x$, the interior equation $u''(x_{i})+\kappa^{2}u(x_{i})=f(x_{i})$ is approximated by the three-point stencil
\begin{equation}
  \frac{u_{i-1}-2u_{i}+u_{i+1}}{\Delta x^{2}}+\kappa^{2}u_{i}=f_{i},
\end{equation}
which places $-2/\Delta x^{2}+\kappa^{2}$ on the main diagonal and $1/\Delta x^{2}$ on the two off-diagonals. The 1D real-valued assembly scales every interior row by $\Delta x^{2}$, an equivalent rescaling that stores the entries $-(2-\kappa^{2}\Delta x^{2})$ and $1$ with the right-hand side $f_{i}\Delta x^{2}$. The two- and three-dimensional problems use the corresponding five- and seven-point Laplacian stencils,
\begin{equation}
\begin{split}
  (\Delta_{h}u)_{ij}={}&
  \frac{u_{i-1,j}-2u_{ij}+u_{i+1,j}}{\Delta x^{2}}\\
  &+\frac{u_{i,j-1}-2u_{ij}+u_{i,j+1}}{\Delta y^{2}},
\end{split}
\end{equation}
together with its seven-point analogue along the third axis, and shift the diagonal entry by $\kappa^{2}$. The assembly admits anisotropic spacing $\Delta x\neq\Delta y\neq\Delta z$, but all reported runs use isotropic grids.

\paragraph{Boundary conditions.}
Dirichlet conditions are imposed in one of two algebraically equivalent ways. In the 1D real, 2D complex and 3D complex problems each boundary node is retained in the system and assigned the identity equation $u=g$, with the prescribed value placed on the right-hand side. These three problems use homogeneous data $u|_{\partial\Omega}=0$, so $g=0$. In the inhomogeneous 1D complex problem the boundary nodes are instead eliminated from the system. Their values $u(0)=1$ and $u(1)=0$ are folded into the interior source vector through the $1/\Delta x^{2}$ coupling.

\paragraph{Complex-valued cases.}
Three of the four benchmarks are assembled directly over $\mathbb{C}$. The 1D complex problem uses squared wavenumber $\kappa^{2}=1-\mathrm{i}$, obtained from $\rho=1$, $\omega=2$ and complex shear modulus $G=2+2\mathrm{i}$ through $\kappa^{2}=\rho\omega^{2}/(G'+\mathrm{i}G'')$, with constant forcing $f=2+3\mathrm{i}$. The 2D and 3D complex problems use $\kappa^{2}=0.8-0.4\mathrm{i}$, obtained from $\rho=\omega=1$ and $G=1+0.5\mathrm{i}$, with constant forcing $f=2+0\mathrm{i}$. Each complex assembly is the finite-difference counterpart of the real2 realification used by LinPDE-GP (Sec.~\ref{subsec:real2}): solving the system over $\mathbb{C}$ is algebraically equivalent to solving the coupled real block system for $(u_{R},u_{I})$. No small-imaginary-part thresholding is applied to these baselines, so the full complex coefficients are retained throughout assembly and solution.

\paragraph{Linear solvers.}
The 1D real and 2D complex systems are assembled in sparse format and solved with a sparse direct solver, and the 1D complex tridiagonal system is solved by dense direct factorization. The 3D complex system on the $161^{3}$ production grid contains $4{,}173{,}281$ unknowns, for which sparse direct factorization is impractical. It is therefore solved by fast diagonalization: a type-I discrete sine transform (DST-I) diagonalizes the seven-point Laplacian under homogeneous Dirichlet data and reduces the solve to $\mathcal{O}(N\log N)$ operations on the identical discrete operator. On a $21^{3}$ grid the fast solver and the sparse direct factorization agree to a maximum modulus of $1.3\times10^{-15}$, which is machine precision for this problem. Because every solver is direct or spectral, the reported errors reflect discretization rather than iterative truncation.

\paragraph{Reference solutions and error evaluation.}
Where a closed-form solution exists, it serves as ground truth. The 1D real problem uses $u^{*}(x)=2\bigl(1-\cos x/\cos 1\bigr)$, and the 1D complex problem uses $u^{*}(x)=c_1\sin(\kappa x)+c_2\cos(\kappa x)+f/\kappa^{2}$, with $c_1$ and $c_2$ fixed by the boundary data. The 2D and 3D constant-forcing problems have no elementary closed form. The reference is therefore the truncated sine eigenfunction expansion of Eq.~\eqref{eqn:nd_analytic}, in which the complex $\kappa^{2}$ enters through the modal denominators $(-\lambda_{\mathbf{k}}+\kappa^{2})^{-1}$. We evaluate it at $d=2$ and $d=3$, retaining odd modes up to index $99$ per direction (fifty modes per axis) in two dimensions and index $49$ per direction (twenty-five modes per axis) in three dimensions. Errors for the 1D problems are evaluated on a uniform test grid of $N_{\text{test}}=1000$ points, onto which the finite-difference solution is linearly interpolated. The 2D and 3D errors are evaluated on the shared $31^{2}$ and $21^{3}$ grids of Sec.~\ref{sec:exp_n_results}; the finite-difference solutions are computed on finer $161^{2}$ and $161^{3}$ grids and mapped onto the evaluation grids by bicubic-spline and trilinear interpolation respectively, applied separately to the real and imaginary parts. The $21^{3}$ evaluation nodes form a subset of the $161^{3}$ solver grid, so the trilinear map is exact there up to rounding. All complex-valued metrics are reported separately for the real and imaginary components.

\paragraph{Convergence.}
Grid-refinement studies confirm the expected accuracy of each discretization. The estimated convergence rate is $2.00$ for the 1D real problem (maximum error) and $1.99$ and $1.97$ for the 2D and 3D complex problems (relative $\ell^{2}$ error of the complex field, over grids from $21^{2}$ to $121^{2}$ and from $11^{3}$ to $81^{3}$). All three agree with the theoretical second-order rate $\mathcal{O}(\Delta x^{2})$ of the central scheme. The finest $161^{3}$ grid is excluded from the 3D rate fit because its error is limited by the truncation of the reference series rather than by the discretization. These studies validate the implementation independently of the comparison with LinPDE-GP.
\begin{table*}[t]
  \caption{\label{tab:fdm_settings}Finite-difference baseline configurations for the four synthetic Helmholtz benchmarks of Sec.~\ref{sec:exp_n_results}. All grids are uniform and isotropic; $\Delta x$ denotes the grid spacing and $N_{\text{fdm}}$ the total node count. The squared wavenumber follows $\kappa^{2}=\rho\omega^{2}/(G'+\mathrm{i}G'')$, and $g$ denotes the prescribed Dirichlet data on $\partial\Omega$. The 1D, 2D and 3D complex systems are assembled and solved directly over $\mathbb{C}$; the 2D and 3D solutions are interpolated from the $161^{2}$ and $161^{3}$ solver grids onto the shared evaluation grids.}
  \begin{ruledtabular}
  \begin{tabular}{lllll}
     & 1D real & 1D complex & 2D complex & 3D complex \\
    \hline
    Domain $\Omega$               & $[-1,1]$            & $[0,1]$                  & $[-1,1]^{2}$          & $[-1,1]^{3}$ \\
    Field $u$                     & real scalar         & complex                  & complex               & complex \\
    Nodes $N_{\text{fdm}}$        & $101$               & $1000$                   & $161\times161=25{,}921$ & $161^{3}=4{,}173{,}281$ \\
    Spacing $\Delta x$            & $0.02$              & $\approx1.0\times10^{-3}$ & $0.0125$             & $0.0125$ \\
    $(\rho,\omega,G',G'')$        & $(1,1,1,0)$         & $(1,2,2,2)$              & $(1,1,1,0.5)$         & $(1,1,1,0.5)$ \\
    $\kappa^{2}$                  & $1$                 & $1-\mathrm{i}$           & $0.8-0.4\mathrm{i}$   & $0.8-0.4\mathrm{i}$ \\
    Source $f$                    & $2$                 & $2+3\mathrm{i}$          & $2+0\mathrm{i}$       & $2+0\mathrm{i}$ \\
    Dirichlet data $g$            & $u(\pm1)=0$         & $u(0)=1,\ u(1)=0$        & $u|_{\partial\Omega}=0$ & $u|_{\partial\Omega}=0$ \\
    Laplacian stencil             & 3-point             & 3-point (complex)        & 5-point (complex)     & 7-point (complex) \\
    Linear solver                 & sparse direct       & dense direct             & sparse direct         & DST-I fast diagonalization \\
    Reference $u^{*}$             & closed form         & closed form              & sine--sine series     & triple-sine series \\
    Error grid                    & $N_{\text{test}}=1000$ & $N_{\text{test}}=1000$ & $31\times31$ (interpolated) & $21^{3}$ (interpolated) \\
  \end{tabular}
  \end{ruledtabular}
\end{table*}
 
\subsection{Physics-informed neural network (PINN)}\label{app:pinn}

\begin{table*}[t]
  \caption{\label{tab:pinn_settings}Physics-informed neural-network baseline
    configurations for the four synthetic Helmholtz benchmarks of Sec.~\ref{sec:exp_n_results}. Network architectures are written as layer-width lists, including the input and output layers; the complex networks have two output channels for the real and imaginary parts. $N_{\text{res}}$ is the number of interior PDE residual points, sampled within the domain except for the 1D complex case, which uses fixed uniform-grid anchors. The squared wavenumber follows $\kappa^{2}=\rho\omega^{2}/(G'+\mathrm{i}G'')$. Reference solutions and error grids match the finite-difference baseline of Appendix~\ref{app:fdm}.}
  \begin{ruledtabular}
  \small
  \begin{tabular}{lllll}
     & 1D real & 1D complex & 2D complex & 3D complex \\
    \hline
    Domain $\Omega$            & $[-1,1]$              & $[0,1]$                       & $[-1,1]^{2}$          & $[-1,1]^{3}$ \\
    Field $u$ (outputs)        & real scalar ($1$)    & complex ($2$: Re, Im)         & complex ($2$: Re, Im) & complex ($2$: Re, Im) \\
    Architecture               & $[1,50,50,50,1]$     & $[1,64,64,64,2]$              & $[2,64,64,64,2]$      & $[3,64,64,64,2]$ \\
    Hidden layers $\times$ width & $3\times50$        & $3\times64$                   & $3\times64$           & $3\times64$ \\
    Activation                 & $\tanh$              & $\tanh$                       & $\tanh$               & $\tanh$ \\
    Initialization             & Glorot uniform       & Glorot normal                 & Glorot normal         & Glorot normal \\
    Boundary handling          & soft (endpoints)     & soft (point-set)              & hard transform        & hard transform \\
    Residual points $N_{\text{res}}$ & $400$          & $998$ (fixed)                 & $2000$                & $8000$ \\
    $(\rho,\omega,G',G'')$     & $(1,1,1,0)$          & $(1,2,2,2)$                   & $(1,1,1,0.5)$         & $(1,1,1,0.5)$ \\
    $\kappa^{2}$               & $1$                  & $1-\mathrm{i}$                & $0.8-0.4\mathrm{i}$   & $0.8-0.4\mathrm{i}$ \\
    Source $f$                 & $2$                  & $2+3\mathrm{i}$               & $2+0\mathrm{i}$       & $2+0\mathrm{i}$ \\
    Optimizer                  & Adam                 & Adam $+$ L-BFGS               & Adam $+$ L-BFGS       & Adam $+$ L-BFGS \\
    Iterations                 & $10^{4}$             & $2\times10^{4}$ $+$ L-BFGS    & $2\times10^{4}$ $+$ L-BFGS & $2\times10^{4}$ $+$ L-BFGS \\
    Learning rate $\eta$       & $10^{-3}$            & $10^{-3}$                     & $10^{-3}$             & $10^{-3}$ \\
    Precision                  & float32              & float64                       & float64               & float64 \\
    Back end                   & TensorFlow           & PyTorch                       & PyTorch               & PyTorch \\
    Hardware                   & L4 GPU               & L4 GPU                        & L4 GPU                & T4 GPU \\
    Reference $u^{*}$          & closed form          & closed form                   & sine--sine series     & triple-sine series \\
    Error grid                 & $N_{\text{test}}=1000$ & $N_{\text{test}}=1000$      & $31\times31$ nodes    & $21^{3}$ nodes \\
  \end{tabular}
  \end{ruledtabular}
\end{table*}

The four synthetic benchmarks of Sec.~\ref{sec:exp_n_results} share a single physics-informed neural-network baseline, implemented with DeepXDE~\citep{Lu_2021} (version $1.15.0$) and configured per experiment as summarized in Table~\ref{tab:pinn_settings}. In every case a fully connected feed-forward network approximates the solution field and is trained to minimize a composite loss that combines the squared Helmholtz residual at interior collocation points with the Dirichlet boundary information. The scheme is deterministic and returns a point estimate, which provides the learning-based reference for the posterior mean of LinPDE-GP. Because the network is supervised by far more interior residual points than the $N_{\text{col}}$ collocation sites used by LinPDE-GP, the comparison is read as a constraint-efficiency benchmark rather than a best-achievable-accuracy contest.

\paragraph{Network architecture.}
Each solver uses a multilayer perceptron with $\tanh$ activations throughout. The input dimension matches the spatial dimension. The 1D real problem uses a single output channel, whereas the three complex problems use two output channels, corresponding to the real and imaginary parts of $u$ under the real2 representation of Sec.~\ref{subsec:real2}. The hidden architecture is three layers of $50$ units for the 1D real problem and three layers of $64$ units for each of the complex problems. Weights are initialized with the Glorot scheme, using the uniform variant for the 1D real problem and the normal variant for the three complex problems.

\paragraph{Boundary conditions.}
Dirichlet data enter through one of two strategies. The 1D problems impose the boundary values as soft penalty terms in the loss. The 1D real problem penalizes the squared boundary residual at $100$ sampled boundary points, which in one dimension coincide with the two endpoints $x\in\{-1,1\}$, and the 1D complex problem enforces $u(0)=1$ and $u(1)=0$ through point-set constraints applied separately to the real and imaginary output channels. The 2D and 3D complex problems instead enforce the homogeneous Dirichlet condition exactly through a hard-constraint output transform, $\hat{u}(\mathbf{x})=\big(\prod_{i}(1-x_{i}^{2})\big)\,\mathrm{NN}_{\theta}(\mathbf{x})$, applied to both output channels of the raw network $\mathrm{NN}_{\theta}$. This transform removes the need for a boundary loss term, so for these two problems the loss reduces to the PDE residuals alone.

\paragraph{Collocation and loss.}
The PDE residual is the squared violation of the Helmholtz equation, $\lvert(\Delta+\kappa^{2})u-f\rvert^{2}$, evaluated at the interior collocation points. For the three complex problems the operator is split into the coupled real system of Sec.~\ref{subsec:real2}: writing $\kappa^{2}=a+\mathrm{i}b$ and $f=f_{\mathrm{re}}+\mathrm{i}f_{\mathrm{im}}$, the residuals $r_{\mathrm{re}}=\Delta u_{\mathrm{re}}+a\,u_{\mathrm{re}}-b\,u_{\mathrm{im}}-f_{\mathrm{re}}$ and $r_{\mathrm{im}}=\Delta u_{\mathrm{im}}+a\,u_{\mathrm{im}}+b\,u_{\mathrm{re}}-f_{\mathrm{im}}$ are minimized jointly. The interior residual count $N_{\text{res}}$ is $400$ for the 1D real problem, $2000$ for the 2D complex problem, and $8000$ for the 3D complex problem, sampled within the domain; held-out sets of $1000$, $1000$, and $4000$ residual points monitor convergence for the 1D real, 2D, and 3D problems. The 1D complex problem uses $998$ fixed anchor points on a uniform grid over $[0,1]$, with no stochastic resampling. The forcing is $f=2$ for the 1D real problem, $f=2+3\mathrm{i}$ for the 1D complex problem, and $f=2+0\mathrm{i}$ for the 2D and 3D complex problems. The squared wavenumber follows $\kappa^{2}=\rho\omega^{2}/(G'+\mathrm{i}G'')$ from the material parameters listed in Table~\ref{tab:pinn_settings}.

\paragraph{Optimization and precision.}
Training uses the Adam optimizer with a fixed learning rate $\eta=10^{-3}$ and no learning rate scheduling. The 1D real problem trains for $10^{4}$ iterations with Adam only. The three complex problems use a two-phase schedule: Adam for $2\times10^{4}$ iterations followed by L-BFGS with DeepXDE's default settings until its internal stopping criterion. Floating-point precision differs by problem. The 1D real problem uses the DeepXDE default single precision (float32), whereas the three complex problems set double precision (float64) for the coupled systems. Random seeds are fixed for reproducibility, using $42$ for the 1D real problem and $0$ for the three complex problems.

\paragraph{Reference solutions and error evaluation.}
The PINN baseline is scored against the same reference solutions as the finite-difference baseline of Appendix~\ref{app:fdm}. The 1D real problem uses the closed form $u^{*}(x)=2\bigl(1-\cos x/\cos 1\bigr)$, and the 1D complex problem uses $u^{*}(x)=c_1\sin(\kappa x)+c_2\cos(\kappa x)+f/\kappa^{2}$, with $c_1$ and $c_2$ fixed by the boundary data. The 2D and 3D complex problems use the sine eigenfunction expansion of Eq.~\eqref{eqn:nd_analytic}, evaluated at $d=2$ and $d=3$ respectively. As in Appendix~\ref{app:fdm}, the series is truncated to odd modes up to index $99$ per direction in two dimensions and index $49$ per direction in three dimensions. Errors for the 1D problems are computed on a uniform test grid of $N_{\text{test}}=1000$ points. The 2D and 3D errors are evaluated at the $31^{2}$ and $21^{3}$ grid nodes, and all complex metrics are reported separately for the real and imaginary components. Because the trained network is a continuous map, it is queried directly at the evaluation points and no interpolation enters the reported metrics.

\paragraph{Implementation.}
The four solvers were run across two DeepXDE back ends, chosen for memory and runtime convenience rather than for any difference in the governing physics. The 1D real problem uses the TensorFlow back end, while the three complex problems use the PyTorch back end. The 1D real, 1D complex and 2D complex problems were run on an L4 (high-RAM) GPU, and the 3D complex problem on a T4 (high-RAM) GPU. Wall-clock training times were approximately $1.2\times10^{3}$\,s for the 2D complex problem and $3.4\times10^{3}$\,s for the 3D complex problem. The governing equations, network definitions, and loss functions are identical across back ends, so the back-end choice does not affect the reported configurations.

\nocite{*}
\bibliography{references}

@article{larsson2003helmholtz,
  title={Helmholtz and parabolic equation solutions to a benchmark problem in ocean acoustics},
  author={Larsson, Elisabeth and Abrahamsson, Leif},
  journal={The Journal of the Acoustical Society of America},
  volume={113},
  number={5},
  pages={2446--2454},
  year={2003},
  publisher={Acoustical Society of America}
}

@book{gumerov2005fast,
  title={Fast multipole methods for the Helmholtz equation in three dimensions},
  author={Gumerov, Nail A and Duraiswami, Ramani},
  year={2005},
  publisher={Elsevier}
}

@book{Rasmussen2006Gaussian,
  author = {Rasmussen, Carl Edward and Williams, Christopher K. I.},
  publisher = {The MIT Press},
  title = {Gaussian Processes for Machine Learning},
  year = 2006
}

@misc{pförtner2024physicsinformedgaussianprocessregression,
      title={Physics-Informed Gaussian Process Regression Generalizes Linear PDE Solvers}, 
      author={Marvin Pförtner and Ingo Steinwart and Philipp Hennig and Jonathan Wenger},
      year={2024},
      eprint={2212.12474},
      archivePrefix={arXiv},
      primaryClass={cs.LG},
      url={https://arxiv.org/abs/2212.12474}, 
}

@book{meyers2008mechanical,
  title={Mechanical behavior of materials},
  author={Meyers, Marc Andr{\'e} and Chawla, Krishan Kumar},
  year={2008},
  publisher={Cambridge university press}
}

@book{bishop2006pattern,
  title={Pattern recognition and machine learning},
  author={Bishop, Christopher M and Nasrabadi, Nasser M},
  volume={4},
  number={4},
  year={2006},
  publisher={Springer}
}

@article{neeser2002proper,
  title={Proper complex random processes with applications to information theory},
  author={Neeser, Fredy D and Massey, James L},
  journal={IEEE transactions on information theory},
  volume={39},
  number={4},
  pages={1293--1302},
  year={2002},
  publisher={IEEE}
}

@book{schreier2010statistical,
  title={Statistical signal processing of complex-valued data: the theory of improper and noncircular signals},
  author={Schreier, Peter J and Scharf, Louis L},
  year={2010},
  publisher={Cambridge university press}
}

@ARTICLE{8307269,
  author={Boloix-Tortosa, Rafael and Murillo-Fuentes, Juan José and Payán-Somet, Francisco Javier and Pérez-Cruz, Fernando},
  journal={IEEE Transactions on Neural Networks and Learning Systems}, 
  title={Complex Gaussian Processes for Regression}, 
  year={2018},
  volume={29},
  number={11},
  pages={5499-5511},
  doi={10.1109/TNNLS.2018.2805019}}

@article{GRYAZIN2003122,
title = {Two numerical methods for an inverse problem for the 2-D Helmholtz equation},
journal = {Journal of Computational Physics},
volume = {184},
number = {1},
pages = {122-148},
year = {2003},
issn = {0021-9991},
doi = {https://doi.org/10.1016/S0021-9991(02)00023-2},
url = {https://www.sciencedirect.com/science/article/pii/S0021999102000232},
author = {Yuriy A. Gryazin and Michael V. Klibanov and Thomas R. Lucas}
}

@book{lahaye2017modern,
  title={Modern solvers for Helmholtz problems},
  author={Lahaye, Domenico and Tang, Jok and Vuik, Kees and others},
  year={2017},
  publisher={Springer}
}

@article{ihlenburg1995finite,
  title={Finite element solution of the Helmholtz equation with high wave number Part I: The h-version of the FEM},
  author={Ihlenburg, Frank and Babu{\v{s}}ka, Ivo},
  journal={Computers \& Mathematics with Applications},
  volume={30},
  number={9},
  pages={9--37},
  year={1995},
  publisher={Elsevier}
}

@article{li2024optimized,
  title={An optimized CIP-FEM to reduce the pollution errors for the Helmholtz equation on a general unstructured mesh},
  author={Li, Buyang and Li, Yonglin and Yang, Zongze},
  journal={Journal of Computational Physics},
  volume={511},
  pages={113120},
  year={2024},
  publisher={Elsevier}
}

@article{erlangga2008advances,
  title={Advances in iterative methods and preconditioners for the Helmholtz equation},
  author={Erlangga, Yogi A},
  journal={Archives of Computational Methods in Engineering},
  volume={15},
  number={1},
  pages={37--66},
  year={2008},
  publisher={Springer}
}

@article{cheng2006wideband,
  title={A wideband fast multipole method for the Helmholtz equation in three dimensions},
  author={Cheng, Hongwei and Crutchfield, William Y and Gimbutas, Zydrunas and Greengard, Leslie F and Ethridge, J Frank and Huang, Jingfang and Rokhlin, Vladimir and Yarvin, Norman and Zhao, Junsheng},
  journal={Journal of Computational Physics},
  volume={216},
  number={1},
  pages={300--325},
  year={2006},
  publisher={Elsevier}
}

@article{martinsson2005fast,
  title={A fast direct solver for boundary integral equations in two dimensions},
  author={Martinsson, Per-Gunnar and Rokhlin, Vladimir},
  journal={Journal of Computational Physics},
  volume={205},
  number={1},
  pages={1--23},
  year={2005},
  publisher={Elsevier}
}

@article{zhao2018time,
  title={Time-domain analysis of power law attenuation in space-fractional wave equations},
  author={Zhao, Xiaofeng and McGough, Robert J},
  journal={The Journal of the Acoustical Society of America},
  volume={144},
  number={1},
  pages={467--477},
  year={2018},
  publisher={AIP Publishing}
}

@article{holm2014comparison,
  title={Comparison of fractional wave equations for power law attenuation in ultrasound and elastography},
  author={Holm, Sverre and N{\"a}sholm, Sven Peter},
  journal={Ultrasound in medicine \& biology},
  volume={40},
  number={4},
  pages={695--703},
  year={2014},
  publisher={Elsevier}
}

@article{carcione1988wave,
  title={Wave propagation simulation in a linear viscoelastic medium},
  author={Carcione, Jos{\'e} M and Kosloff, Dan and Kosloff, Ronnie},
  journal={Geophysical Journal International},
  volume={95},
  number={3},
  pages={597--611},
  year={1988},
  publisher={Blackwell Publishing Ltd Oxford, UK}
}

@article{kjartansson1979constant,
  title={Constant Q-wave propagation and attenuation},
  author={Kjartansson, Einar},
  journal={Journal of Geophysical Research: Solid Earth},
  volume={84},
  number={B9},
  pages={4737--4748},
  year={1979},
  publisher={Wiley Online Library}
}

@article{karniadakis2021physics,
  title={Physics-informed machine learning},
  author={Karniadakis, George Em and Kevrekidis, Ioannis G and Lu, Lu and Perdikaris, Paris and Wang, Sifan and Yang, Liu},
  journal={Nature Reviews Physics},
  volume={3},
  number={6},
  pages={422--440},
  year={2021},
  publisher={Nature Publishing Group UK London}
}

@article{cai2021physics,
  title={Physics-informed neural networks (PINNs) for fluid mechanics: A review},
  author={Cai, Shengze and Mao, Zhiping and Wang, Zhicheng and Yin, Minglang and Karniadakis, George Em},
  journal={Acta Mechanica Sinica},
  volume={37},
  number={12},
  pages={1727--1738},
  year={2021},
  publisher={Springer}
}

@article{raissi2019physics,
  title={Physics-informed neural networks: A deep learning framework for solving forward and inverse problems involving nonlinear partial differential equations},
  author={Raissi, Maziar and Perdikaris, Paris and Karniadakis, George E},
  journal={Journal of Computational physics},
  volume={378},
  pages={686--707},
  year={2019},
  publisher={Elsevier}
}

@article{yang2021b,
  title={B-PINNs: Bayesian physics-informed neural networks for forward and inverse PDE problems with noisy data},
  author={Yang, Liu and Meng, Xuhui and Karniadakis, George Em},
  journal={Journal of Computational Physics},
  volume={425},
  pages={109913},
  year={2021},
  publisher={Elsevier}
}

@article{cuomo2022scientific,
  title={Scientific machine learning through physics--informed neural networks: Where we are and what’s next},
  author={Cuomo, Salvatore and Di Cola, Vincenzo Schiano and Giampaolo, Fabio and Rozza, Gianluigi and Raissi, Maziar and Piccialli, Francesco},
  journal={Journal of Scientific Computing},
  volume={92},
  number={3},
  pages={88},
  year={2022},
  publisher={Springer}
}

@article{hennig2015probabilistic,
  title={Probabilistic numerics and uncertainty in computations},
  author={Hennig, Philipp and Osborne, Michael A and Girolami, Mark},
  journal={Proceedings of the Royal Society A: Mathematical, Physical and Engineering Sciences},
  volume={471},
  number={2179},
  pages={20150142},
  year={2015},
  publisher={The Royal Society Publishing}
}

@inproceedings{cockayne2017probabilistic,
  title={Probabilistic numerical methods for PDE-constrained Bayesian inverse problems},
  author={Cockayne, Jon and Oates, Chris and Sullivan, Tim and Girolami, Mark},
  booktitle={AIP Conference Proceedings},
  volume={1853},
  number={1},
  pages={060001},
  year={2017},
  organization={AIP Publishing LLC}
}

@inproceedings{albert2019gaussian,
  title={Gaussian processes for data fulfilling linear differential equations},
  author={Albert, Christopher G},
  booktitle={Proceedings},
  volume={33},
  number={1},
  pages={5},
  year={2019},
  organization={MDPI}
}

@article{ambrogioni2019complex,
  title={Complex-valued gaussian process regression for time series analysis},
  author={Ambrogioni, Luca and Maris, Eric},
  journal={Signal Processing},
  volume={160},
  pages={215--228},
  year={2019},
  publisher={Elsevier}
}

@article{boloix2018complex,
  title={Complex Gaussian processes for regression},
  author={Boloix-Tortosa, Rafael and Murillo-Fuentes, Juan Jos{\'e} and Pay{\'a}n-Somet, Francisco Javier and P{\'e}rez-Cruz, Fernando},
  journal={IEEE transactions on neural networks and learning systems},
  volume={29},
  number={11},
  pages={5499--5511},
  year={2018},
  publisher={IEEE}
}

@INPROCEEDINGS{7178363,
  author={Tobar, Felipe and Turner, Richard E.},
  booktitle={2015 IEEE International Conference on Acoustics, Speech and Signal Processing (ICASSP)}, 
  title={Modelling of complex signals using gaussian processes}, 
  year={2015},
  volume={},
  number={},
  pages={2209-2213},
  doi={10.1109/ICASSP.2015.7178363}
}

@article{THOMASSEALE20161781,
title = {The simulation of magnetic resonance elastography through atherosclerosis},
journal = {Journal of Biomechanics},
volume = {49},
number = {9},
pages = {1781-1788},
year = {2016},
issn = {0021-9290},
doi = {https://doi.org/10.1016/j.jbiomech.2016.04.013},
url = {https://www.sciencedirect.com/science/article/pii/S0021929016304651},
author = {L.E.J. Thomas-Seale and L. Hollis and D. Klatt and I. Sack and N. Roberts and P. Pankaj and P.R. Hoskins}
}

@article{perreard2010effects,
  title={Effects of frequency-and direction-dependent elastic materials on linearly elastic MRE image reconstructions},
  author={Perreard, IM and Pattison, AJ and Doyley, M and McGarry, MDJ and Barani, Z and Van Houten, EE and Weaver, JB and Paulsen, KD},
  journal={Physics in Medicine \& Biology},
  volume={55},
  number={22},
  pages={6801--6815},
  year={2010}
}

@article{bayly2021mr,
  title={MR imaging of human brain mechanics in vivo: new measurements to facilitate the development of computational models of brain injury},
  author={Bayly, Philip V and Alshareef, Ahmed and Knutsen, Andrew K and Upadhyay, Kshitiz and Okamoto, Ruth J and Carass, Aaron and Butman, John A and Pham, Dzung L and Prince, Jerry L and Ramesh, KT and others},
  journal={Annals of biomedical engineering},
  volume={49},
  number={10},
  pages={2677--2692},
  year={2021},
  publisher={Springer}
}

@article{upadhyay2022development,
  title={Development and validation of subject-specific 3D human head models based on a nonlinear visco-hyperelastic constitutive framework},
  author={Upadhyay, Kshitiz and Alshareef, Ahmed and Knutsen, Andrew K and Johnson, Curtis L and Carass, Aaron and Bayly, Philip V and Pham, Dzung L and Prince, Jerry L and Ramesh, KT},
  journal={Journal of the Royal Society Interface},
  volume={19},
  number={195},
  pages={20220561},
  year={2022}
}

@book{goovaerts1997geostatistics,
  title={Geostatistics for natural resources evaluation},
  author={Goovaerts, Pierre},
  year={1997},
  publisher={Oxford university press}
}

@article{alvarez2012kernels,
  title={Kernels for vector-valued functions: A review},
  author={Alvarez, Mauricio A and Rosasco, Lorenzo and Lawrence, Neil D},
  journal={Foundations and Trends{\textregistered} in Machine Learning},
  volume={4},
  number={3},
  pages={195--266},
  year={2012},
  publisher={Emerald Publishing Limited}
}

@article{journel1976mining,
  title={Mining geostatistics},
  author={Journel, Andre G and Huijbregts, Charles J},
  year={1976}
}

@article{murphy2013measuring,
  title={Measuring the characteristic topography of brain stiffness with magnetic resonance elastography},
  author={Murphy, Matthew C and Huston III, John and Jack Jr, Clifford R and Glaser, Kevin J and Senjem, Matthew L and Chen, Jun and Manduca, Armando and Felmlee, Joel P and Ehman, Richard L},
  journal={PloS one},
  volume={8},
  number={12},
  pages={e81668},
  year={2013},
  publisher={Public Library of Science San Francisco, USA}
}

@article{oehlert1992note,
  title={A note on the delta method},
  author={Oehlert, Gary W},
  journal={The American Statistician},
  volume={46},
  number={1},
  pages={27--29},
  year={1992},
  publisher={Taylor \& Francis}
}

@article{Lu_2021,
   title={DeepXDE: A Deep Learning Library for Solving Differential Equations},
   volume={63},
   ISSN={1095-7200},
   url={http://dx.doi.org/10.1137/19M1274067},
   DOI={10.1137/19m1274067},
   number={1},
   journal={SIAM Review},
   publisher={Society for Industrial & Applied Mathematics (SIAM)},
   author={Lu, Lu and Meng, Xuhui and Mao, Zhiping and Karniadakis, George Em},
   year={2021},
   month=Jan, pages={208–228} }

@article{babuvska1995generalized,
  title={A generalized finite element method for solving the Helmholtz equation in two dimensions with minimal pollution},
  author={Babu{\v{s}}ka, Ivo and Ihlenburg, Frank and Paik, Ellen T and Sauter, Stefan A},
  journal={Computer methods in applied mechanics and engineering},
  volume={128},
  number={3-4},
  pages={325--359},
  year={1995},
  publisher={Elsevier}
}

@article{farhat2000two,
  title={Two-level domain decomposition methods with Lagrange multipliers for the fast iterative solution of acoustic scattering problems},
  author={Farhat, Charbel and Macedo, Antonini and Lesoinne, Michel and Roux, Francois-Xavier and Magoul{\'e}s, Fr{\'e}d{\'e}ric and de La Bourdonnaie, Armel},
  journal={Computer methods in applied mechanics and engineering},
  volume={184},
  number={2-4},
  pages={213--239},
  year={2000},
  publisher={Elsevier}
}

@article{diwan2021iterative,
  title={Iterative solution with shifted Laplace preconditioner for plane wave enriched isogeometric analysis and finite element discretization for high-frequency acoustics},
  author={Diwan, Ganesh C and Mohamed, M Shadi},
  journal={Computer Methods in Applied Mechanics and Engineering},
  volume={384},
  pages={114006},
  year={2021},
  publisher={Elsevier}
}

@article{mao2020physics,
  title={Physics-informed neural networks for high-speed flows},
  author={Mao, Zhiping and Jagtap, Ameya D and Karniadakis, George Em},
  journal={Computer Methods in Applied Mechanics and Engineering},
  volume={360},
  pages={112789},
  year={2020},
  publisher={Elsevier}
}

@article{jagtap2020conservative,
  title={Conservative physics-informed neural networks on discrete domains for conservation laws: Applications to forward and inverse problems},
  author={Jagtap, Ameya D and Kharazmi, Ehsan and Karniadakis, George Em},
  journal={Computer Methods in Applied Mechanics and Engineering},
  volume={365},
  pages={113028},
  year={2020},
  publisher={Elsevier}
}

@article{kharazmi2021hp,
  title={hp-VPINNs: Variational physics-informed neural networks with domain decomposition},
  author={Kharazmi, Ehsan and Zhang, Zhongqiang and Karniadakis, George Em},
  journal={Computer Methods in Applied Mechanics and Engineering},
  volume={374},
  pages={113547},
  year={2021},
  publisher={Elsevier}
}

@article{linka2022bayesian,
  title={Bayesian physics informed neural networks for real-world nonlinear dynamical systems},
  author={Linka, Kevin and Sch{\"a}fer, Amelie and Meng, Xuhui and Zou, Zongren and Karniadakis, George Em and Kuhl, Ellen},
  journal={Computer Methods in Applied Mechanics and Engineering},
  volume={402},
  pages={115346},
  year={2022},
  publisher={Elsevier}
}

@article{zou2025uncertainty,
  title={Uncertainty quantification for noisy inputs--outputs in physics-informed neural networks and neural operators},
  author={Zou, Zongren and Meng, Xuhui and Karniadakis, George Em},
  journal={Computer Methods in Applied Mechanics and Engineering},
  volume={433},
  pages={117479},
  year={2025},
  publisher={Elsevier}
}

@article{abdulle2021probabilistic,
  title={A probabilistic finite element method based on random meshes: A posteriori error estimators and Bayesian inverse problems},
  author={Abdulle, Assyr and Garegnani, Giacomo},
  journal={Computer Methods in Applied Mechanics and Engineering},
  volume={384},
  pages={113961},
  year={2021},
  publisher={Elsevier}
}

@article{nguyen2015gaussian,
  title={Gaussian functional regression for linear partial differential equations},
  author={Nguyen, Ngoc Cuong and Peraire, Jaime},
  journal={Computer Methods in Applied Mechanics and Engineering},
  volume={287},
  pages={69--89},
  year={2015},
  publisher={Elsevier}
}

@article{gulian2022gaussian,
  title={Gaussian process regression constrained by boundary value problems},
  author={Gulian, Mamikon and Frankel, Ari and Swiler, Laura},
  journal={Computer Methods in Applied Mechanics and Engineering},
  volume={388},
  pages={114117},
  year={2022},
  publisher={Elsevier}
}

@article{girolami2021statistical,
  title={The statistical finite element method (statFEM) for coherent synthesis of observation data and model predictions},
  author={Girolami, Mark and Febrianto, Eky and Yin, Ge and Cirak, Fehmi},
  journal={Computer Methods in Applied Mechanics and Engineering},
  volume={375},
  pages={113533},
  year={2021},
  publisher={Elsevier}
}

@article{dalton2026finite,
  title={Finite-element Gaussian processes for the machine learning of steady-state linear partial differential equations},
  author={Dalton, David and Gao, Hao and Husmeier, Dirk},
  journal={Computer Methods in Applied Mechanics and Engineering},
  volume={451},
  pages={118580},
  year={2026},
  publisher={Elsevier}
}

@article{mao2025decomposition,
  title={Decomposition-enhanced parameterized physics-informed neural network for uncertainty-aware interior acoustic analysis},
  author={Mao, Runpeng and Qiang, Xin and Wang, Chong},
  journal={Computer Methods in Applied Mechanics and Engineering},
  volume={446},
  pages={118279},
  year={2025},
  publisher={Elsevier}
}

@article{zou2026probabilistic,
  title={A probabilistic framework for learning solution operators of deterministic high-frequency Helmholtz equations},
  author={Zou, Yicheng and Lanthaler, Samuel and Salahshoor, Hossein},
  journal={Computer Methods in Applied Mechanics and Engineering},
  volume={461},
  pages={119160},
  year={2026},
  publisher={Elsevier}
}

@article{upadhyay2022data,
  title={Data-driven uncertainty quantification in computational human head models},
  author={Upadhyay, Kshitiz and Giovanis, Dimitris G and Alshareef, Ahmed and Knutsen, Andrew K and Johnson, Curtis L and Carass, Aaron and Bayly, Philip V and Shields, Michael D and Ramesh, KT},
  journal={Computer methods in applied mechanics and engineering},
  volume={398},
  pages={115108},
  year={2022},
  publisher={Elsevier}
}

\end{document}